\documentclass[manuscript,screen]{acmart}
\AtBeginDocument{%
  }

\setcopyright{none}
\copyrightyear{2025}
\acmYear{2025}
\acmDOI{XXXXXXX.XXXXXXX}

\acmConference[Conference acronym 'XX]{Make sure to enter the correct
  conference title from your rights confirmation emai}{June 03--05,
  2018}{Woodstock, NY}
\acmISBN{978-1-4503-XXXX-X/18/06}

\usepackage{float}
\usepackage{longtable}
\usepackage{graphicx}
\usepackage{lineno} 
\begin{document}

\title{Video Is Worth a Thousand Images: Exploring the Latest Trends in Long Video Generation}

\author{Faraz Waseem}
\email{faraz.waseem@pgr.reading.ac.uk}
\orcid{1234-5678-9012}
\author{Muhammad Shahzad}
\authornotemark[1]
\email{m.shahzad2@reading.ac.uk}
\affiliation{%
  \institution{University Of Reading}
  \city{Reading}
  \state{Berkshire}
  \country{UK}
}

\renewcommand{\shortauthors}{Faraz and Shahzad}

\begin{abstract}

An image may convey a thousand words, but a video, composed of hundreds or thousands of image frames, tells a more intricate story. Despite significant progress in multimodal large language models (MLLMs), generating extended videos remains a formidable challenge. As of this writing, OpenAI’s Sora~\cite{sora}, the current state-of-the-art system, is still limited to producing videos of up to one minute in length. This limitation stems from the complexity of long video generation, which requires more than generative AI techniques for approximating density functions. Critical elements, such as planning, narrative construction, and spatiotemporal continuity, pose significant challenges. Integrating generative AI with a divide-and-conquer approach could improve scalability for longer videos while offering greater control. In this survey, we examine the current landscape of long video generation, covering foundational techniques such as GANs and diffusion models, video generation strategies, large-scale training datasets, quality metrics for evaluating long videos, and future research areas to address the limitations of existing video generation capabilities. We believe it would serve as a comprehensive foundation, offering extensive information to guide future advancements and research in the field of long video generation.

\end{abstract}




\maketitle









\section{Introduction}  
The year 2022 marked a significant milestone in the field of the generative AI era with the release of ChatGPT~\cite{ChatGPT}. ChatGPT is an advanced language model that produces human-like text from user input, supporting tasks like answering questions, creative
writing, and conversation. This technology uses complex deep neural networks based on large language models trained
on extensive text data to capture intricate linguistic patterns and contextual nuances for precise text generation and
understanding. Since then, major tech companies have introduced their large language models (LLMS), such as Facebook's
LLama series~\cite{llama}, Google's Gemini~\cite{gemini}, and a few other notable models, including Claude~\cite{Claude} and Mistral~\cite{mistral}.

The success of LLMs brought a transformative breakthrough in image generation. DALL-E 2~\cite{ramesh2022hierarchicaltextconditionalimagegeneration} surpassed traditional GANs and VAEs by interpreting natural language and generating diverse concepts and styles, excelling in photorealistic outputs. Other systems, such as Stable Diffusion 3~\cite{esser2024scalingrectifiedflowtransformers} and MidJourney~\cite{midjourney}, also demonstrate strong capabilities in creating realistic visuals.

\begin{figure}[t]
    \centering
    \includegraphics[scale=0.3]{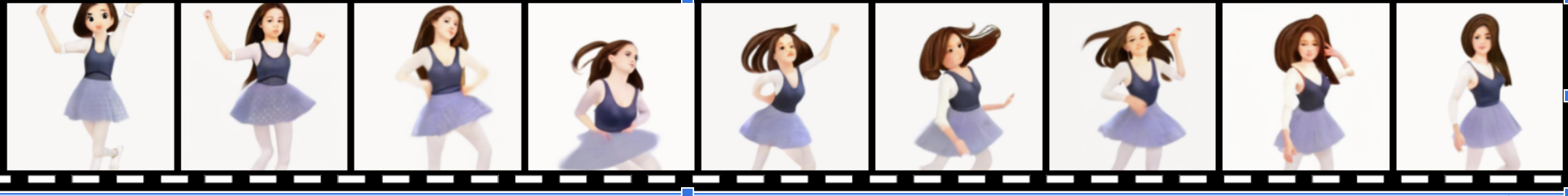}
    \caption[Example of semantic content not changing with the progress of frames~\cite{hong2022cogvideolargescalepretrainingtexttovideo}.]{Example of semantic content not changing with the progress of frames~\cite{hong2022cogvideolargescalepretrainingtexttovideo}.}
    \label{static frames}
\end{figure}

\begin{figure}[t]
    \centering
    \includegraphics[scale=0.3]{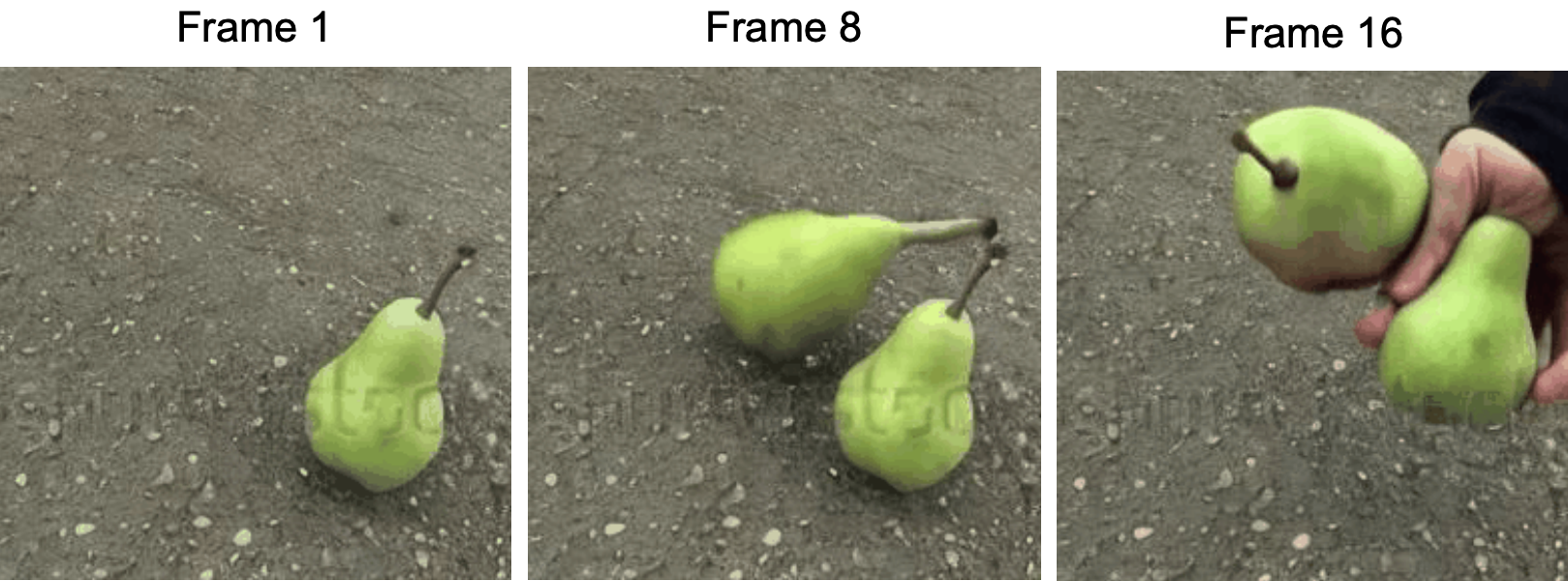}
    \caption[Example of semantic content changing with the progress of frames~\cite{lin2024videodirectorgpt}.]{Example Of semantic content changing with the progress of frames~\cite{lin2024videodirectorgpt}.}
    \label{dynamic frames}
\end{figure}

Video generation is far more complex than text or images due to dynamic elements like motion, occlusion, and evolving semantic content—the conceptual mapping of objects, actions, and interactions. Single-scene videos (e.g., a girl dancing against a static background, Fig. \ref{static frames}) maintain consistent semantics, while multi-scene videos (Fig. \ref{dynamic frames}) introduce new objects or actors, altering semantic content over time.

\begin{figure}[H]
    \centering
    \includegraphics[scale=0.25]{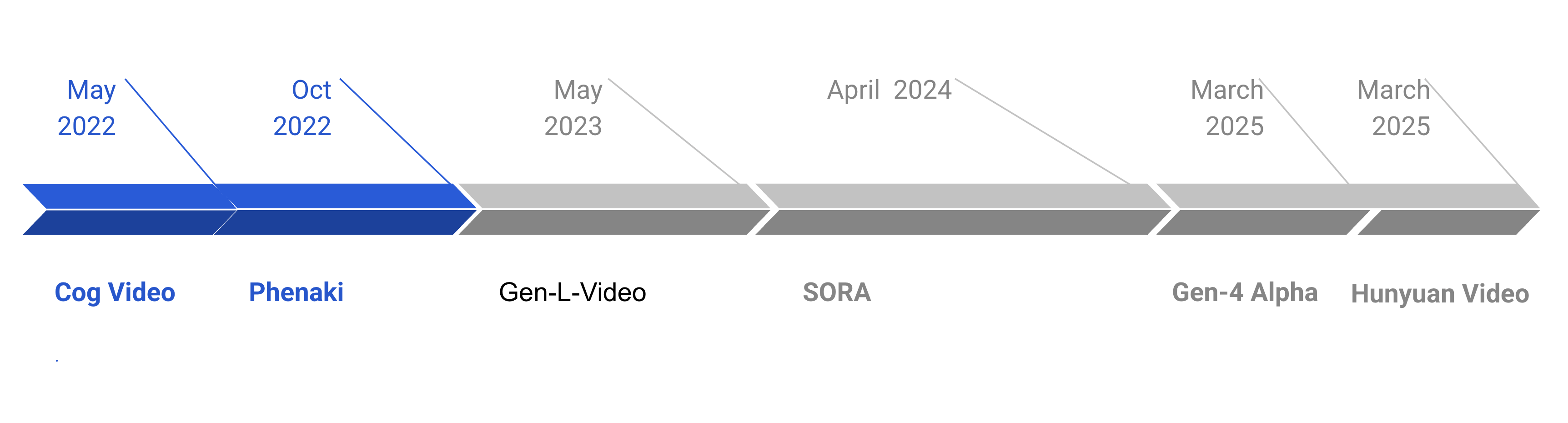}
    \caption{Evolution of long video generation models. Later models, such as SORA~\cite{sora} and Gen-4 Alpha~\cite{runwalmlaplha3}, focus more on video quality than duration.}
    \label{papers timeline}
\end{figure}

Due to the complexities of dynamic scenes, early video generation models were limited to producing short clips lasting only a few seconds, often animating a single static frame without incorporating varying backgrounds or objects. For example,  Make-A-Video ~\cite{singer2022makeavideotexttovideogenerationtextvideo} and  RunwayML Gen-2 ~\cite{runwaymlgen2} generate 4-5 second videos using a single animated frame with little change in semantic content. CogVideo ~\cite{hong2022cogvideo} is among the first long video generation models to create extended videos using autoregressive transformers. However, it operated based on a single prompt and also exhibited minimal changes in the semantic content of the video. Phenaki \cite{villegas2022phenakivariablelengthvideo}, which employs autoregressive video transformers, is one of the first models to generate long videos with dynamic semantic content based on multiple prompts. Similarly, Gen-L-Video ~\cite{wang2023genlvideomultitextlongvideo} employs a diffusion model to merge short video clips into a seamless, continuous video. Sora ~\cite{sora} has established a new state-of-the-art in video generation. Sora is also known as the ChatGPT of the video generation world. 
It utilizes diffusion transformers and sampling from a compressed spatiotemporal representation, producing videos that are visually impressive and rich in semantic content. Another state-of-the-art model in long video generation is Gen-4 Alpha by RunwayML ~\cite{runwalmlaplha3}. It is a commercial model that can generate photo-realistic videos up to 10 seconds long and is also based on diffusion transformers. Fig. \ref{papers timeline} illustrates the evolution of these long video generation techniques, which are still in their early stages compared to human-made, commercial videos, particularly in terms of video length and narrative organization.

\begin{figure}[H]
    \centering
    \includegraphics[scale=0.4]{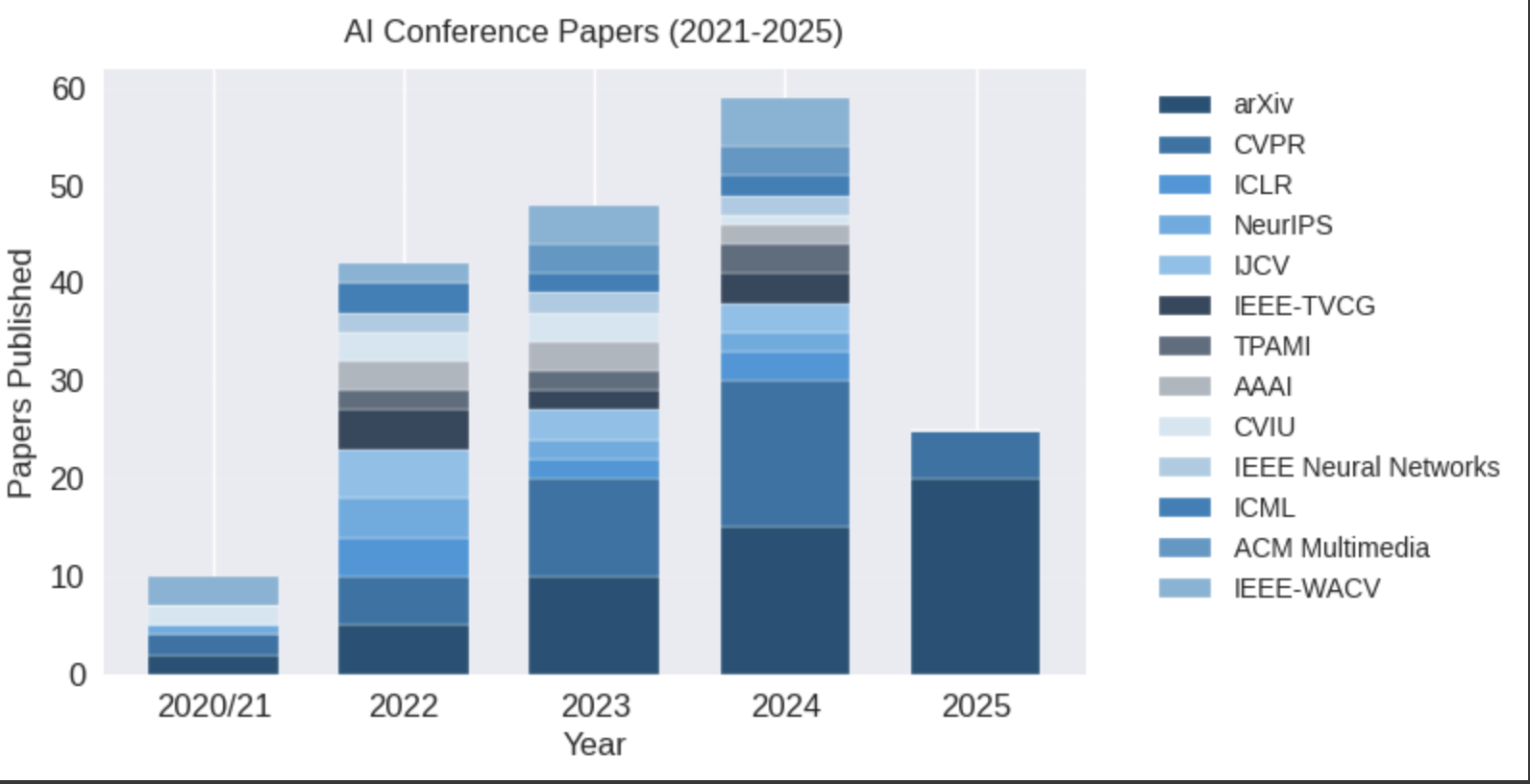}
    \caption{Most papers focusing on long video generation were published in 2023--2025.}
    \label{Publications}
\end{figure}

\subsection{Survey Contributions — Need for Summarizing Long Video Generation Methods}

Long video generation presents numerous challenges beyond crafting and maintaining consistent storylines across scenes. These include the lack of large-scale video datasets with detailed captions and the requirement of significant computational resources. Despite these challenges, long video generation has emerged as a transformative frontier in generative AI, unlocking novel possibilities in various fields, including entertainment, education, healthcare, marketing, and gaming. This potential has sparked substantial research attention, and publication rates have accelerated dramatically in recent years. As illustrated in Figure \ref{Publications}, the field has experienced explosive growth, with more than two-thirds of all academic work produced within the past 24 months—a testament to its dynamic transformation.

Given these interests and opportunities, it is time to summarize the state-of-the-art in long video generation and discuss the associated challenges, progress, and future directions that will support the advancement of the field. To our knowledge, there are only two related long video-generation surveys  
~\cite{li2024surveylongvideogeneration} and ~\cite{zhou2024surveygenerativeaillm}. The former~\cite{li2024surveylongvideogeneration} explores the latest trends in long video generation, highlighting the divide-and-conquer and autoregressive approaches as two primary themes. It also examines photo-realism trends and generative paradigms, such as VAE, GAN, and diffusion-based models. However, while it highlights the divide-and-conquer approach, which simplifies the complexity of long videos by breaking them into smaller, manageable chunks, a detailed exploration of this methodology, such as how short videos can be seamlessly integrated into longer narratives, is lacking. Our work aims to bridge this gap by thoroughly analyzing the various dimensions of the divide-and-conquer strategy and its role in addressing the challenges of long video generation. In contrast, the latter~\cite{zhou2024surveygenerativeaillm} provides a broader overview of the video generation field, encompassing topics such as long videos, video editing, super-resolution, datasets, and metrics. However, long-video generation is only one of many topics discussed. It highlights the need for a focused and in-depth study on the generation of long videos. Our contribution addresses this need by providing a comprehensive analysis of this emerging field and highlighting key approaches, challenges, and prospects.

The proposed work addresses the gaps in the existing literature by incorporating recent studies and conducting a comprehensive analysis of the divide-and-conquer approach. Specifically, we focus on underexplored aspects, such as agent-based networks and methods for transitioning from short to long videos, which are missing from current reviews. Furthermore, we extend our examination to methodologies and papers that fall outside the divide-and-conquer and autoregressive categories, as well as the latest research not covered by previous surveys. In addition, we dive deeply into the literature, emphasizing the algorithms, models, and input control techniques employed in generating long videos, offering a more holistic and detailed perspective on the field.

\subsection{Survey Focus — Techniques, Challenges \& Key Questions in Long Video Generation}

Video generation utilizes various techniques, such as sampling from latent space~\cite{liu2024sorareviewbackgroundtechnology}, creating small video segments or images, generating intermediate frames ("divide and conquer")[\ref{sec:Divide-Conquer-Paradigms}], employing autoregressive methods [\ref{sec:autoregressive-methods}] to predict future frames based on initial ones, and improving latent state representations for longer videos. Training video generation models presents challenges due to the higher computational requirements and more extensive memory needs for video datasets. Many video generation models are based on pretrained image models [~\cite{huang2023freebloom}, ~\cite{lu2023flowzero}, ~\cite{blattmann2023alignlatentshighresolutionvideo}], which enhance attention mechanisms to ensure consistency between adjacent frames, as video is essentially a sequence of frames. Some long video models are developed by extending short video generation models[~\cite{zhang2024mavinmultiactionvideogeneration},~\cite{chen2023seineshorttolongvideodiffusion},~\cite{wang2023genlvideomultitextlongvideo}] and improving control mechanisms for longer content. Another significant aspect of video generation is input guidance[[\ref{sec:inputTextPrompt}],[\ref{sec:inputTextPromptWithScenesAndImages}],[\ref{sec:inputTextPromptWithScenes}] ]. Long video generation requires stronger guidance than images or short clips, typically anchored in text embeddings, such as CLIP~\cite{radford2021learningtransferablevisualmodels}. Here, large language models (Section~\ref{sec:llmasdirector}) take center stage: they decode physical world dynamics, forecast object interactions, and choreograph multistep actions, leveraging their pre-trained knowledge to steer generation toward coherent, long-form outputs. When evaluating the quality of the generated videos[[\ref{sec:imagequality}],[\ref{sec:videoquality}],[~\ref{sec:semanticsquality}],[~\ref{sec:compositequality}]], it is important to assess the quality of the individual frames, the fluidity of motion and the overall aesthetic appeal. Ensuring that generated videos remain faithful to the input text while preserving entity consistency (e.g., cars, actors) across frames is a critical challenge. This has motivated researchers to explore novel directions in the field, raising key questions that warrant further investigation.

\begin{enumerate}
\item How can we generate long videos with multiple semantic segments with different actors, actions, and objects? 
\item How can we ensure semantic consistency across long video segments, such as maintaining consistent models of objects like cars?
\item  Discussion of Long-video generation strategies covering segmented stitching, auto-regressive frames, and full latent-space synthesis.

\end{enumerate}

Our survey article centers around these critical questions, providing insights to guide researchers and practitioners in addressing these challenges.

\subsection{Survey Methodology} 

For this survey, we conducted searches across several conferences, including but not limited to CVPR, ICLR, NeurIPS, IJCV, IEEE-TVCG, TPAMI, AAAI, CVIU, IEEE Neural Networks, ICML, ACM Multimedia, IEEE-WACV. We used keywords such as "video generation," "long video generation," and "LLM-guided video generation." Additionally, we searched academic databases, including arXiv, Google Scholar, IEEE Xplore, ACM Transactions, and Scopus, focusing on the term "long video generation" for our survey. Our survey covers papers published between 2021 and 2025 (as of May 2025), with a focus on "video generation" and "generative AI". We gathered over 190+ articles through snowball sampling, using keywords such as text-to-video, generative AI, visual interpretation, and extended video generation.

\subsection{Survey Organization}

 We will begin by discussing the \hyperref[sec:videogenfoundation]{foundational frameworks} for video generation, including embedding and LLMs, to set the stage for more advanced topics. The goal is to familiarize readers with these fundamental components, enabling them to explore these building blocks according to their level of expertise and interest.
Next, we will explore \hyperref[{sec:video-gen-paradigm}]{backbone mechanisms} for video generation, including divide-and-conquer autoregressive models and the use of implicit latent spaces. We will then explore \hyperref[{sec:tokenization}]{Tokenization Strategies}. We will explore \hyperref[{sec:video-input-control}]{input guidance mechanisms}, including strategies such as LLM guidance, and categorize them into different levels based on the depth of control the LLM exerts over them. We will also address the necessary modifications to the image and video diffusion models to facilitate such control. We will also discuss the post-processing pipelines required to achieve high temporal and spatial quality in videos generated by diffusion models. We will then discuss the datasets used to train video generation models, as outlined in Section  \hyperref[{sec:data}]{datasets} . We will then discuss the metrics used to measure generated video quality, as outlined in Section \hyperref[{sec:metrics}]{metrics}. Finally, we will talk about future trends and open challenges.

\section{Long Video Generation: Backbone Architectures and Methods}   ~\label{sec:videogenfoundation}
Progress in long-video generation builds on advancements in many foundational building blocks. These include GANs-based architecture \ref{sec:ganvideogen}, Autoencoders ~\ref{sec:variationalvideo}, Transformers-based models \ref{sec:videotransformer}, LLMs and language understanding \ref{sec:llm}, and Image and Video Diffusion models \ref{sec:diffusionmodelsfamily}. 

\subsection {GAN Based Video Generation}  ~\label{sec:ganvideogen}
GANs \cite{goodfellow2014generativeadversarialnetworks} dominated generative tasks from 2014 until the early 2020s, although diffusion models and transformer-based approaches in terms of performance and versatility have since surpassed them. The fundamental GAN framework \cite{goodfellow2014generativeadversarialnetworks} consists of two competing components: a Generator that creates samples from random noise and a Discriminator that evaluates their authenticity. While this adversarial architecture established the foundation for image and video generation, newer methods have advanced beyond its capabilities, as discussed in subsequent sections.

\subsubsection{GAN Based Image Generation} ~\label{sec:imagegeneration}
GANs initially revolutionized image generation, dominating the field of image generation. Here, we examine key ideas and milestones in GAN literature, organized by timeline.


\textit{Early GANS:} GANs ~\cite{goodfellow2014generativeadversarialnetworks} was the first to generate images using adversarial networks but employed simple feedforward neural networks for both the Generator and the Discriminator. DCGANs ~\cite{radford2016unsupervisedrepresentationlearningdeep} extend the GAN architecture by incorporating convolutional layers, making them more suitable for image data. They generated images with a resolution of 64×64. LAPGAN ~\cite{denton2015deepgenerativeimagemodels} increased the resolution of images by generating them multi-scale. It consists of multiple GANs, each generating images at different resolutions. GANs, DCGANs, and LAPGAN are primarily designed to create images based on random noise vectors. These models lack text guidance, but text-based GANs incorporate text-based control, which we will discuss in the next section.


\textit{GAN Text Input:} 
StackGAN ~\cite{zhang2017stackgantextphotorealisticimage} is a multistage text-to-image GAN that generates high-quality images. It has two GAN stacks stacked on top of each other. The first one takes the text and generates a low-resolution image. The second one takes both text and input images and creates high-quality images. AttnGAN  ~\cite{xu2017attnganfinegrainedtextimage} uses attention mechanisms to create images from text, allowing it to focus on specific words or phrases in the input description.    

\textit{Style/Image Transfer:} StyleGAN \cite{karras2020analyzingimprovingimagequality} was the first generative model to generate high-quality artistic images, and one of the key innovations was transferring an artistic style like Vincent Van Gogh's to real-world pictures and image translation. CycleGAN ~\cite{zhu2020unpairedimagetoimagetranslationusing} does image-to-image translation and consists of two generators and two discriminators. StyleGAN2 ~\cite{viazovetskyi2020stylegan2distillationfeedforwardimage} primarily focuses on generating high-quality, diverse images, particularly faces. It introduced disentangled latent space. Latent space is where a vector of N dimensions represents each image. Projecting different high-level attributes, such as skin color and hairstyle, onto distinct dimensions in a latent space provides excellent editing capabilities for realistic image generation, semantic manipulation, and local editing. StyleGAN2\cite{viazovetskyi2020stylegan2distillationfeedforwardimage} opened the doors for high-level image manipulation. StyleGAN2 \cite{viazovetskyi2020stylegan2distillationfeedforwardimage} is an improvement over StyleGAN, producing higher-quality images. pix2pix ~\cite{isola2018imagetoimagetranslationconditionaladversarial} specifically designed for image-to-image translation tasks.  It learns a conditional generative model and generates an output image conditioned on the input image. GAN also revolutionized video generation, which we will explore in ~\ref{sec:ganvideogan}.

\subsubsection { Video/Multi Frame Generation} ~\label{sec:ganvideogan}

\textit{Early Attempts:}~\cite{mathieu2016deepmultiscalevideoprediction}, which generated future frames from observed sequences. ~\cite{vondrick2016generatingvideosscenedynamics} advanced this by using separate 2D and 3D convolutional networks for static backgrounds and moving foregrounds, producing 32-frame unconditional videos of various scenes. Further development came with ~\cite{8578349}'s two-stage model, which first generated 128×128 resolution time-lapse videos from a single frame and then enhanced them with dynamic motion information. These early works laid the important foundations for unconditional video generation before the advent of prompt-based approaches.

\textit{Prompt Based Guidance:} Numerous studies have explored the use of conditional inputs in GAN to guide and refine the generation process. These conditions can take various forms, including audio signals, text prompts, semantic maps, images, or other videos. TGANs-C ~\cite{pan2018createtellgeneratingvideos} incorporate text guidance using LSTM-based latent vectors. TGANs-C was designed to input a single sentence. 

\textit{Long Video Generation Using GAN:} DIGAN ~\cite{yu2022generatingvideosdynamicsawareimplicit} can create 128 frame video. It introduced an INR (Implicit Neural Representations) based video generator that improves motion dynamics by manipulating space and time coordinates differently and a motion discriminator that efficiently identifies unnatural motions without requiring long frame sequences. StyleGan-V ~\cite{skorokhodov2022styleganvcontinuousvideogenerator} improved the state of the art and was built on StyleGAN2 ~\cite{viazovetskyi2020stylegan2distillationfeedforwardimage}. It can generate high-resolution 1024-long videos by designing a holistic discriminator that aggregates temporal information by simply concatenating frame features, thereby decreasing the training cost. 

\subsection {Autoencoder Based Video Generation} ~\label{sec:variationalvideo}

Autoencoders ~\cite{autoencoder}, variational autoencoders ~\cite{kingma2022autoencodingvariationalbayes}, and masked autoencoders ~\cite{he2021maskedautoencodersscalablevision} belong to the family of models that compress information into a compact latent space and serve as building blocks for image and video generation pipelines. Masked autoencoders can generate videos from this learned latent space. We will discuss the foundations of autoencoders and build up the video generation process via masked autoencoders.

\subsubsection {Autoencoder Formulation} ~\label{sec:encoderformulation}  Autoencoder is an unsupervised neural network
that compresses its input into a compact latent layer and then learns to reproduce its input through backpropagation. The autoencoder is trained to minimize the reconstruction loss between the input \( \mathbf{x} \) and the reconstructed output \( \mathbf{\hat{x}} \). 
The most common application of an autoencoder for long video and video generation is the construction of compressed latent space. For example, in \cite{rombach2022highresolutionimagesynthesislatent}, the authors use an encoder and decoder to project images from pixel space to latent space, thereby decreasing the computational complexity of learning the image distribution. 

Variational Autoencoders (VAE) ~\cite{kingma2022autoencodingvariationalbayes} address the limitations of traditional autoencoders by learning latent distributions instead of fixed representations, allowing new data generation. The VQ-VAE variant ~\cite{oord2018neuraldiscreterepresentationlearning} has become foundational for video/image generation pipelines like VideoGen ~\cite{yan2021videogptvideogenerationusing}, VQGAN ~\cite{esser2021tamingtransformershighresolutionimage}, and DALL-E ~\cite{ramesh2021zeroshottexttoimagegeneration}. Hybrid approaches like Hierarchical Patch VAE-GAN ~\cite{gur2020hierarchicalpatchvaegangenerating} combine VAEs with GANs, while applications extend to anomaly detection through architectures like LSTM-Convolutional VAEs ~\cite{waseem2022visualanomalydetectionvideo}.

\subsubsection {Masked Autoencoders }~\label{sec:masked autoencoders}

Masked autoencoders ~\cite{he2021maskedautoencodersscalablevision} serve as scalable self-supervised backbones for video generation by reconstructing randomly masked image patches. The approach extends to video through models like VideoMAC ~\cite{pei2024videomacvideomaskedautoencoders}, which uses convolutional networks to reconstruct symmetrically masked frame pairs at high masking ratios (0.75). The framework has evolved into advanced variants, such as MAGVIT ~\cite{Yu_2023_CVPR}, which achieves significantly faster inference than diffusion models, and MAGVLT ~\cite{kim2023magvltmaskedgenerativevisionandlanguage}, which unifies vision-language generation under this paradigm.

\subsection{ Transformer Based Video Generation}~\label{sec:videotransformer}

GANs have limitations, such as mode collapse \cite{hoang2017multigeneratorgenerativeadversarialnets}, training instability, which requires fine-tuning of parameters, and a significant amount of training time and resources. Transformers, introduced in 2017 ~\cite{vaswani2023attentionneed}, made inroads into image and generation via autoregressive and masked encoding. Some of the key concepts to understand are Vision Transformers and Video Transformers.

\subsubsection { Transformer Based Image Generation}~\label{sec:imagegantransformer}

Vision Transformers (ViT) ~\cite{dosovitskiy2021imageworth16x16words} revolutionized computer vision by processing images as patch-based tokens, similar to NLP transformers. DALL-E ~\cite{pmlr-v139-ramesh21a} pioneered this approach for image generation, using a Discrete VAE ~\cite{rolfe2017discretevariationalautoencoders} to compress images into 32×32 tokens and training a GPT-style transformer on 250M image-text pairs. CogView ~\cite{ding2021cogviewmasteringtexttoimagegeneration} later surpassed DALL-E in FID scores but maintained weaker complex prompt rendering. Both autoregressive models suffered from slow generation due to token-by-token processing, a limitation addressed by CogView2 ~\cite{ding2022cogview2fasterbettertexttoimage} through masked cross-modal training.

\subsubsection{ Autoregressive  Based Video Generation} ~\label{sec:imagetovideotransformer}

Video transformers (VViT) \cite{arnab2021vivitvideovisiontransformer} extend vision transformers \cite{dosovitskiy2021imageworth16x16words} by tokenizing video patches. Phenaki \cite{villegas2022phenaki} generates long videos from text prompts using T5X embeddings \cite{roberts2022scalingmodelsdatatextttt5x} and C-ViViT, a variant of VViT \cite{arnab2021vivitvideovisiontransformer} that compresses tokens and employs masked and autoregressive prediction for long sequences. CogVideo \cite{hong2022cogvideo} builds on CogView2 \cite{ding2022cogview2fasterbettertexttoimage}, using hierarchical training for better text-video alignment and a two-stage process involving keyframe generation and interpolation.

\subsection{Language Understanding In Video Generation} ~\label{sec:llm}

\subsubsection {Text To Image Feature Representation}  ~\label{sec:clipimagegen}

The core principle behind text-based visual generation tasks is effectively pairing text with the visual content. Many visual generation pipelines leverage pre-existing image-text pair models, such as CLIP (Contrastive Language-Image Pretraining) ~\cite{radford2021learningtransferablevisualmodels}. CLIP has been pre-trained using a contrastive learning approach that optimizes the cosine similarity between image and text embedding. Given CLIP's robust performance, many visual generation models, such as DALL·E 2 ~\cite{ramesh2022hierarchicaltextconditionalimagegeneration}, incorporate CLIP’s text embedding to leverage its superior semantic understanding. It allows these models to enhance their ability to generate visually relevant and contextually appropriate content, effectively bridging the gap between text and visual representation.

\subsubsection {LLMs Based Video Guidance} ~\label{sec:llmvideogan}

Many visual generation models, such as LLM Director ~\cite{zhu2024compositional3dawarevideogeneration}, leverage standalone large language models (LLMs) ~\cite{brown2020languagemodelsfewshotlearners} to enhance their performance. By integrating LLM, visual generation models can benefit from advanced natural language processing capabilities, enabling them to interpret and generate more nuanced and contextually relevant descriptions of captions in single or multiple prompts, along with detailed scenes. One example of this design is LLM grounded VDM ~\cite{lian2024llmgroundedvideodiffusionmodels}. When paired with visual inputs,  LLMs can transform simple image descriptions into more elaborate storytelling, adding layers of meaning and context that enhance the viewer's experience. LLM can also act as the director of the entire video generation process and create a coherent script, as shown by Vlogger~\cite{zhuang2024vloggermakedreamvlog}. The details on how the recent long video generation methods leverage LLMs are explained in \ref{sec:llmasdirector}.

\subsection { Diffusion Models} ~\label{sec:diffusionmodelsfamily}

Diffusion models have emerged as the state-of-the-art approach for image and video generation, combining components such as variational autoencoders, transformers, and language models. The foundational work ~\cite{sohldickstein2015deepunsupervisedlearningusing} established key principles by applying non-equilibrium thermodynamics to unsupervised learning, while ~\cite{ho2020denoisingdiffusionprobabilisticmodels} advanced the field through parameterized Markov chains trained by variational inference. These breakthroughs created the basis for modern diffusion architectures in visual generation tasks.

\subsubsection{ Image Diffusion } ~\label{sec:difusionsconcepts}

Image diffusion models generate images through iterative denoising, with ~\cite{song2020generativemodelingestimatinggradients} contributing gradient-based estimation methods and ~\cite{ho2020denoisingdiffusionprobabilisticmodels} establishing the foundational DDPM framework. For a text-conditioned generation, models typically employ a U-Net with cross-attention layers using embeddings from CLIP ~\cite{radford2021learningtransferablevisualmodels}, BERT ~\cite{devlin2019bertpretrainingdeepbidirectional}, or T5 ~\cite{raffel2023exploringlimitstransferlearning}. Robin Rombach et al. implemented this in~\cite{rombach2022highresolutionimagesynthesislatent} through modified attention layers that combine multimodal embeddings ~\cite{vaswani2023attentionneed}.

\subsubsection {Video Generation From Diffusion models}~\label{sec:videogandifusion}

Video generation models primarily use two architectures: 3D U-Nets and Transformers. The U-Net approach extends 2D diffusion models to handle 4D tensors (frames × height × width × channels) through factorized spatial-temporal attention, where spatial attention focuses on intra-frame regions and temporal attention captures inter-frame dependencies. Alternatively, Sora ~\cite{soraworldmodel} implements a Transformer-based diffusion model ~\cite{peebles2023scalablediffusionmodelstransformers} that processes videos as space-time patch tokens, as detailed in ~\cite{soraworldmodel}. These architectural approaches for long-video generation are further explored in subsequent sections.

\section{Long Video: Generation Paradigm} ~\label{sec:video-gen-paradigm}

We summarize various video generation approaches into three core paradigms.

\begin{itemize}
  \item Auto-Regressive Paradigm: Videos are generated sequentially, with each frame conditioned on the frames generated before.
  \item Divide-and-conquer approach: Videos are produced by creating keyframes or short video segments guided by storyline prompts, often with the aid of a large language model (LLM). 
  \item Implicit Video Generation: Videos are generated implicitly from the model without needing explicit extrapolation (autoregressive approach) or explicit interpolation (divide-and-conquer) by designing latent space to represent variable-size videos. 
 
\end{itemize}

These approaches will be explained in the following sections.

\subsection{Auto Regressive Approaches } ~\label{sec:autoregressive-methods}

\begin{figure}[t]
    \centering
    \includegraphics[scale=0.25]{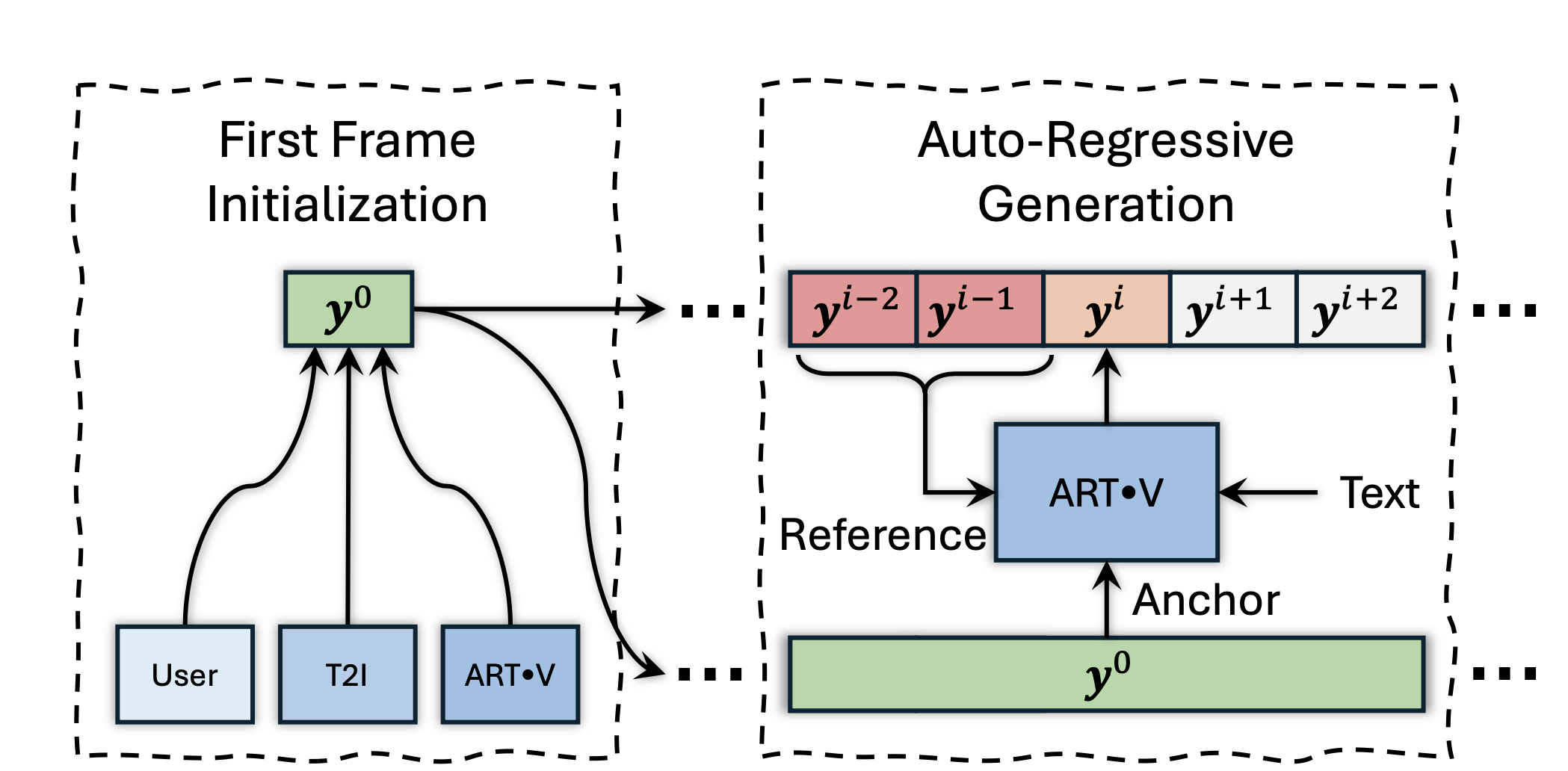}
    \caption{The basic theme of the auto-regressive approach is that it generates new frames, given the initial anchor frame, previous frames, and optional progressive prompts~\cite{Weng_2024_CVPR}.}
    \label{Autoregressive-gen}
\end{figure}

The autoregressive generation paradigm creates videos by sequentially predicting future frames from previous ones, ensuring temporal coherence through this frame-by-frame approach [~\cite{ding2021cogview},~\cite{villegas2022phenakivariablelengthvideo},~\cite{tian2024videotetriscompositionaltexttovideogeneration}]. While effective for maintaining consistency (Fig. \ref{Autoregressive-gen}), this method faces computational limitations for long videos due to its sequential nature. Key implementations include CogVideo ~\cite{hong2022cogvideo}, which extends CogView2 ~\cite{ding2022cogview2} but is constrained by sequence length and resolution (160×160, upscalable to 480×480), and NUWA-Infinity ~\cite{wu2022nuwainfinityautoregressiveautoregressivegeneration} which improves resolution through hierarchical generation. Phenaki ~\cite{villegas2022phenakivariablelengthvideo} advances the paradigm by handling multiple prompts through C-ViViT ~\cite{arnab2021vivitvideovisiontransformer} compression and T5X embeddings ~\cite{roberts2022scalingmodelsdatatextttt5x}, though its bidirectional training requires significant memory.

\textcolor{black} {A paradigm shift emerged with VideoPoet~\cite{10.5555/3692070.3693075}, which demonstrated that multimodal training—combining text, images, and audio in a decoder-only transformer—could overcome these quality limitations. By pretraining on diverse objectives and fine-tuning for specific tasks, VideoPoet achieved state-of-the-art zero-shot generation, proving that autoregressive models can produce high-fidelity videos when augmented with rich multimodal signals.}

\begin{figure}[t]
    \centering
    \includegraphics[scale=0.38]{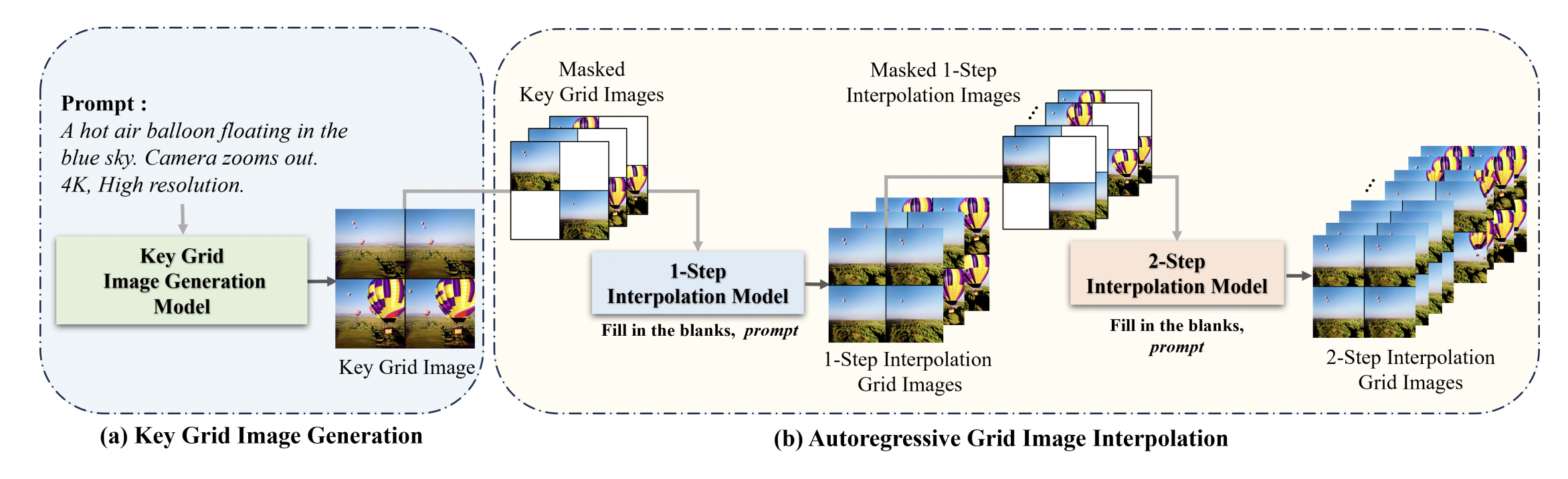}
    \caption { Grid Diffusion Model. It first generates a grid image and then learns a spatial auto-regressive model by learning to predict masked subframe conditions on previous images and unmasked frames.~\cite{lee2024grid}. }
    \label{grid-difusion-design}
\end{figure}

All approaches discussed [~\cite{wu2022nuwainfinityautoregressiveautoregressivegeneration}],[~\cite{ding2021cogview}],[~\cite{villegas2022phenakivariablelengthvideo}] are based on transformers.  Grid Diffusion ~\cite{lee2024grid} is based on diffusion. Grid Diffusion first used compression and represented video using an image created from keyframes, which covers the primary motions or events of the video. It is called a 'grid image,' which consists of 4  subframes representing video keyframes. During the training phase, they masked these frames and learned to produce masked frames conditioned on previous grid images and non-masked images. This design paradigm is illustrated in Fig. \ref{grid-difusion-design}. As they replaced the challenge of video generation with image generation, they can create long videos up to 128 frames autoregressively with high image quality (low FVD scores ~\cite{unterthiner2019accurategenerativemodelsvideo}. They utilized transfer learning from a pre-trained stable diffusion model~\cite{rombach2022highresolutionimagesynthesislatent} and trained on only two Nvidia A100 GPUs. Fig. \ref{grid-difusion-design} explains this architecture. Building on this insight, \textcolor{black} {ARLON ~\cite{li2025arlonboostingdiffusiontransformers} combines the strengths of autoregressive transformers and diffusion models through its Asymmetric Diffusion Transformer (AsymmDiT) and latent VQ-VAE, achieving 128× compression (8× spatial + 6× temporal downsampling) in a 12-channel latent space. This hybrid approach maintains motion fidelity while enabling scalable, high-quality synthesis—effectively bridging diffusion models (Sec.~\ref{sec:diffusionmodelsfamily}) and token-based methods.}

\textcolor{black} {
Long video generation faces critical challenges in memory management and temporal coherence across extended sequences. ARLON ~\cite{li2025arlonboostingdiffusiontransformers} addresses this via a training-free autoregressive inference method using pre-trained diffusion models. Its sliding-window queue mechanism processes frames with progressively increasing noise levels: fully denoised frames are removed from the head of the queue, while new noisy frames are added to the tail. This approach enables the unbounded generation of infinitely long videos without retraining, maintaining computational efficiency through a fixed-size queue that prevents memory overload.
}

Previous methods for video generation, including NUWA-Infinity ~\cite{wu2022nuwainfinityautoregressiveautoregressivegeneration}, CogView ~\cite{ding2021cogview}, Phenaki ~\cite{villegas2022phenakivariablelengthvideo}, and GRID ~\cite{lee2024grid}, face limitations in rendering complex compositional prompts that describe dynamic spatiotemporal interactions—such as "a man walking with a black dog on his right while a blue car drives from the opposite direction." VideoTetris \cite{tian2024videotetriscompositionaltexttovideogeneration} addresses this challenge by introducing spatiotemporal Compositional Diffusion, which manipulates cross-attention maps in denoising networks to synthesize videos that adhere to intricate or evolving instructions. This approach enables the generation of long videos with progressive compositional prompts, where "progressive" refers to continuous changes in object positions, quantities, and attributes, ensuring precise alignment of interacting entities across space and time.

\begin{longtable}{ |c |p{9cm} |c|} 
\caption{Auto Regressive Approaches.}
\label{table:autoregressivecatalog} \\
\hline  
 \textbf{Model}   & \textbf{Theme}  & \textbf{Month/Year} \\ 
 \hline
\endfirsthead
\hline
\textbf{Model}   & \textbf{Theme}  & \textbf{Month/Year}\\ \hline
\endhead

\hline
\endfoot

\hline
\endlastfoot
    
StyleGAN-V~\cite{Skorokhodov_2022_CVPR}&
 Time-continuous signals~\cite{sitzmann2020implicitneuralrepresentationsperiodic}~\cite{mildenhall2020nerfrepresentingscenesneural} extended from StyleGAN2~\cite{viazovetskyi2020stylegan2distillationfeedforwardimage} 
  & Dec 2021   \\
\hline
     
DIGAN~\cite{yu2022generatingvideosdynamicsawareimplicit}&
Implict neural representation based~\cite{sitzmann2020implicitneuralrepresentationsperiodic} video generation model & Feb 2022  \\
\hline      
CogVideo~\cite{hong2022cogvideo}&
Long videos using autoregressive and interpolation stages & May 2022 \\
\hline
NUWA-Infinity~\cite{wu2022nuwainfinityautoregressiveautoregressivegeneration}&
Long videos with hierarchical autoregressive modeling & July 2022  \\
\hline
Phenaki~\cite{villegas2022phenakivariablelengthvideo}&
Compresses videos into discrete tokens for efficient frame generation & Oct 2022 \\
\hline

PVDM ~\cite{yu2023videoprobabilisticdiffusionmodels}&
PVDM uses diffusion in latent space for video generation & Feb 2023 \\
\hline

MeBT ~\cite{yoo2023endtoendgenerativemodelinglong}&
Memory efficient transformer for long-range dependency videos & March 2023  \\
\hline

ART•V~\cite{Weng_2024_CVPR}&
Auto-regressive using keyframes and image diffusion & Nov 2023  \\
\hline

StreamingT2V~\cite{henschel2024streamingt2vconsistentdynamicextendable} &
 Long videos with consistent transitions and scene preservation
& March 2024 \\
\hline

Grid Diffusion Models ~\cite{lee2024grid}&
Video generation by merging four keyframes into images & March 2024  \\
 \hline 

ViD-GPT~\cite{gao2024vidgptintroducinggptstyleautoregressive}&
GPT-style autoregressive generation into video diffusion models & June 2024  \\
\hline

FlexiFilm ~\cite{ouyang2024flexifilmlongvideogeneration} &
Long videos with temporal conditioning and resampling strategy & June 2024  \\
\hline
\textcolor{black} {VideoPoet} ~\cite{10.5555/3692070.3693075} &
 \textcolor{black} {An autoregressive LLM for high-quality synthesis from multi modal inputs } & June 2024 \\
\hline

VideoTetris~\cite{tian2024videotetriscompositionaltexttovideogeneration} &
Text-to-video generation with spatio-temporal compositional diffusion & Oct 2024 \\
\hline

\textcolor{black} {Arlon}~\cite{li2025arlonboostingdiffusiontransformers} &
\textcolor{black} {AR for long-range temporal guidance and DiT for high-fidelity synthesis }& Jan 2025 \\
\hline

\end{longtable}

Autoregressive video generation has progressed from CogVideo ~\cite{ding2021cogview}'s single-prompt, low-resolution outputs to modern systems like VideoTetris, which can model complex scientific dynamics with multiple prompts while preserving quality. While autoregressive methods excel at smooth motion transitions, their sequential nature results in slow generation and limited control over complex scene elements (actors, bounding boxes, spatial relationships). Divide-and-conquer approaches (Sec. \ref{sec:Divide-Conquer-Paradigms}) address these limitations by enabling parallel frame generation and leveraging LLMs for structured video blueprints, improving the handling of dynamic scenes. Key papers and insights are cataloged in Table \ref{table:autoregressivecatalog}.

\subsection{Divide And Conquer Paradigms}  ~\label{sec:Divide-Conquer-Paradigms}

\begin{figure}[t]
    \centering
    \includegraphics[scale=0.3]{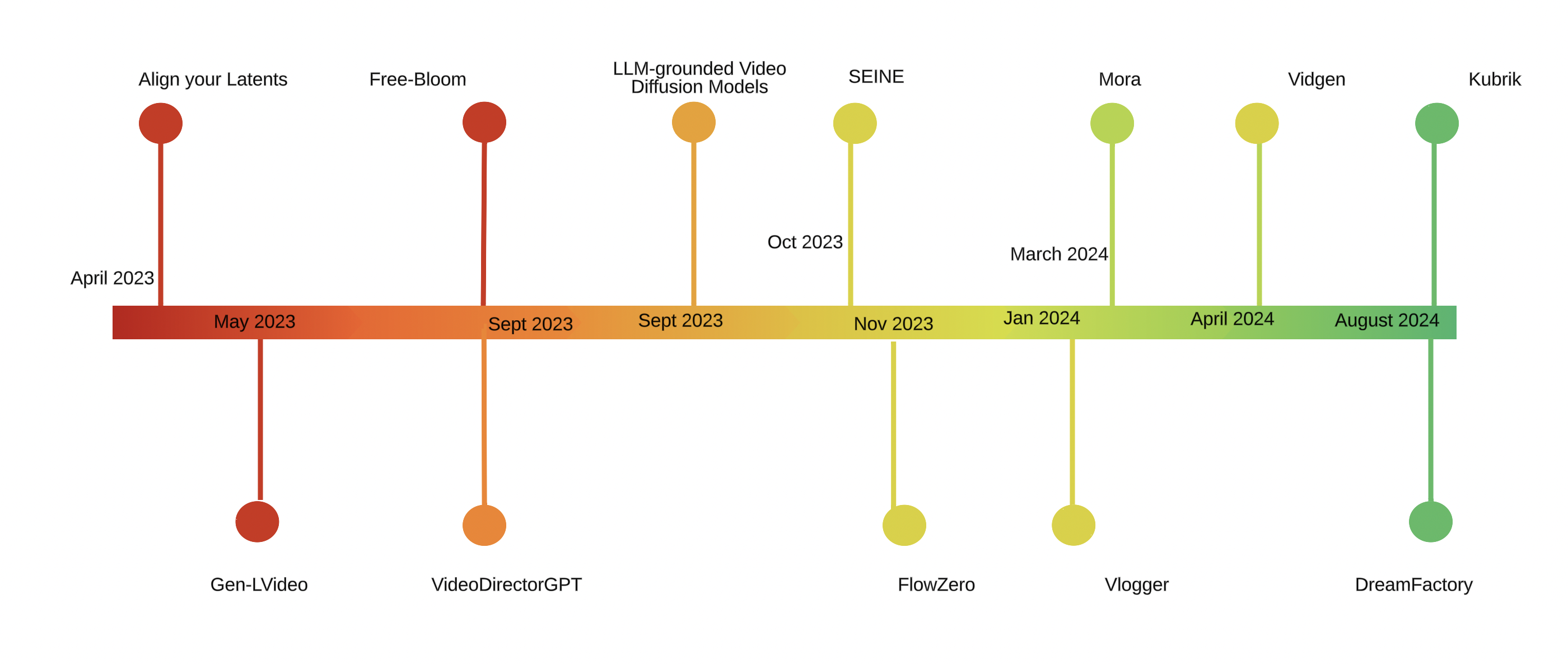}
    \caption{Divide-and-conquer timeline: We used the dates these papers were published in online resources, such as arXiv or https://openreview.net/. Papers catalog here are Align your Latents~\cite{blattmann2023alignlatentshighresolutionvideo}, Gen-L-Video~\cite{wang2023genlvideomultitextlongvideo}, Free-Bloom~\cite{huang2023freebloom}, VideoDirectorGPT~\cite{lin2024videodirectorgpt}, LLM-grounded Video Diffusion Models~\cite{lian2024llmgroundedvideodiffusionmodels}, SEINE~\cite{chen2023seineshorttolongvideodiffusion}, FlowZero~\cite{lu2023flowzero}, Mora~\cite{yuan2024moraenablinggeneralistvideo}, Vlogger~\cite{zhuang2024vloggermakedreamvlog}, Vidgen~\cite{10544050}, DreamFactory~\cite{xie2024dreamfactorypioneeringmultiscenelong} and Kubrik~\cite{he2024kubrickmultimodalagentcollaborations} }
    \label{divide-and-conquer-timeline}
\end{figure}

The basic theme of divide and conquer is to generate keyframes or short clips based on single or multiple prompts and then interpolate between these frames or clips. The system often uses an anchor image as a reference frame for generating all subsequent frames. The system generates each key frame independently, allowing for parallel processing. Some challenges associated with the divide-and-conquer approach include maintaining semantic consistency, ensuring smooth motion transformation between interpolated frames, and achieving high-quality frames. One key theme of the Divide and Conquer approach is separating the planning and video generation stages, differentiating it from the autoregressive approach as illustrated in Fig. \ref{llm-as-director}. Divide and Conquer Paradigms can be divided into three sub-paradigms: the Large Language Model as Director, the Intermediate Transition Model, and the Agent-Based Framework.
Some milestone papers with timelines are illustrated in Fig. \ref{divide-and-conquer-timeline}.

\subsubsection{Large Language Model As Director  } ~\label{sec:llmasdirector}

\begin{figure}[H]
    \centering
    \includegraphics[scale=0.35]{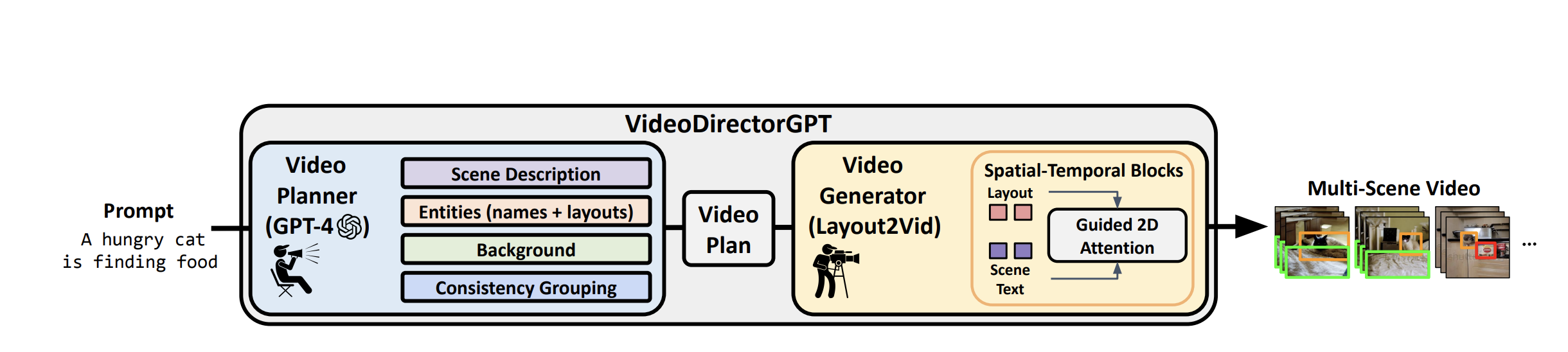}
    \caption{VideoDirectorGPT: GPT-4 generates a blueprint for video generation, including scene and entity description. Separate module Layout2Vid generates video from this video plan~\cite{lin2024videodirectorgpt}}
    \label{Video-Director-Paper}
\end{figure}

The LLM-as-Director paradigm ~\cite{lin2024videodirectorgpt}~\cite{wang2023modelscopetexttovideotechnicalreport}~\cite{10544050}~\cite{huang2023freebloomzeroshottexttovideogenerator} revolutionizes video generation by employing a two-stage process: (1) an LLM Planner creates detailed narrative blueprints (keyframes, layouts, actions) and (2) a Video Generator Backbone produces intermediate frames (Fig. \ref{Video-Director-Paper}). This framework supports both zero-shot (training-free) and training-based approaches.

Free-Bloom ~\cite{huang2023freebloom} demonstrates zero-shot capability through innovative techniques like joint noise sampling and DDIM-based dual-path interpolation ~\cite{song2022denoisingdiffusionimplicitmodels}, though its LLM scripting potential is underutilized. VideoDirectorGPT ~\cite{lin2024videodirectorgpt} enhances this with GPT-4's comprehensive planning (layouts, bounding boxes) executed via Layout2Vid ~\cite{wang2023modelscopetexttovideotechnicalreport}. The training-based LLM-grounded model ~\cite{lian2024llmgrounded} further improves realism by learning spatiotemporal dynamics from the text.
\begin{figure}[t]
    \centering
    \includegraphics[scale=0.5]{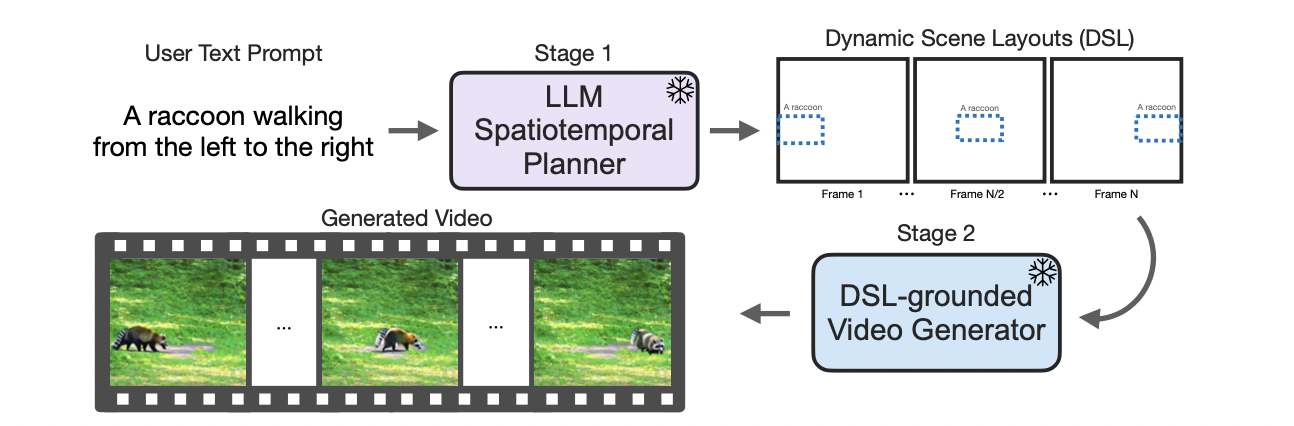}
    \caption{The LLM as director approach utilizes LLM as the spatiotemporal director of the script, along with a separate video generation module that can understand the DSL (metadata) generated by LLM. ~\cite{lian2024llmgroundedvideodiffusionmodels}}
    \label{llm-as-director}
\end{figure}

Like VideoDirectorGPT, FlowZero ~\cite{lu2023flowzero} adopted a zero-shot (training-free) approach; however, the LLM plays a more detailed role than Video Director GPT. It generates a detailed dynamic scene syntax (DSS), including scene descriptions, object arrangements, and background motion patterns. The DSS components direct an image diffusion model to generate videos with smooth object movements and consistent frame transitions. The theme of using dynamic scene layout is illustrated in Fig. \ref{llm-as-director}.  The LLM as director approach has limitations, as it is a two-stage architecture, and adding speech or integrating short clips will require modifications in the pipeline. We can extend the LLM-based divide-and-conquer approach by incorporating more specialized components, such as a model for generating reference images, a module for video creation, and a plug-and-play module for adding transitional clips and speech. That is achieved using the multi-agent framework, as described in the section.
 \ref{section:Agent-Based Divide And Conquer approach}.

\subsubsection{Multi stage/Agent Based  Divide And Conquer Approach  } ~\label{section:Agent-Based Divide And Conquer approach}

\begin{figure}[H]
    \centering
    \includegraphics[scale=0.5]{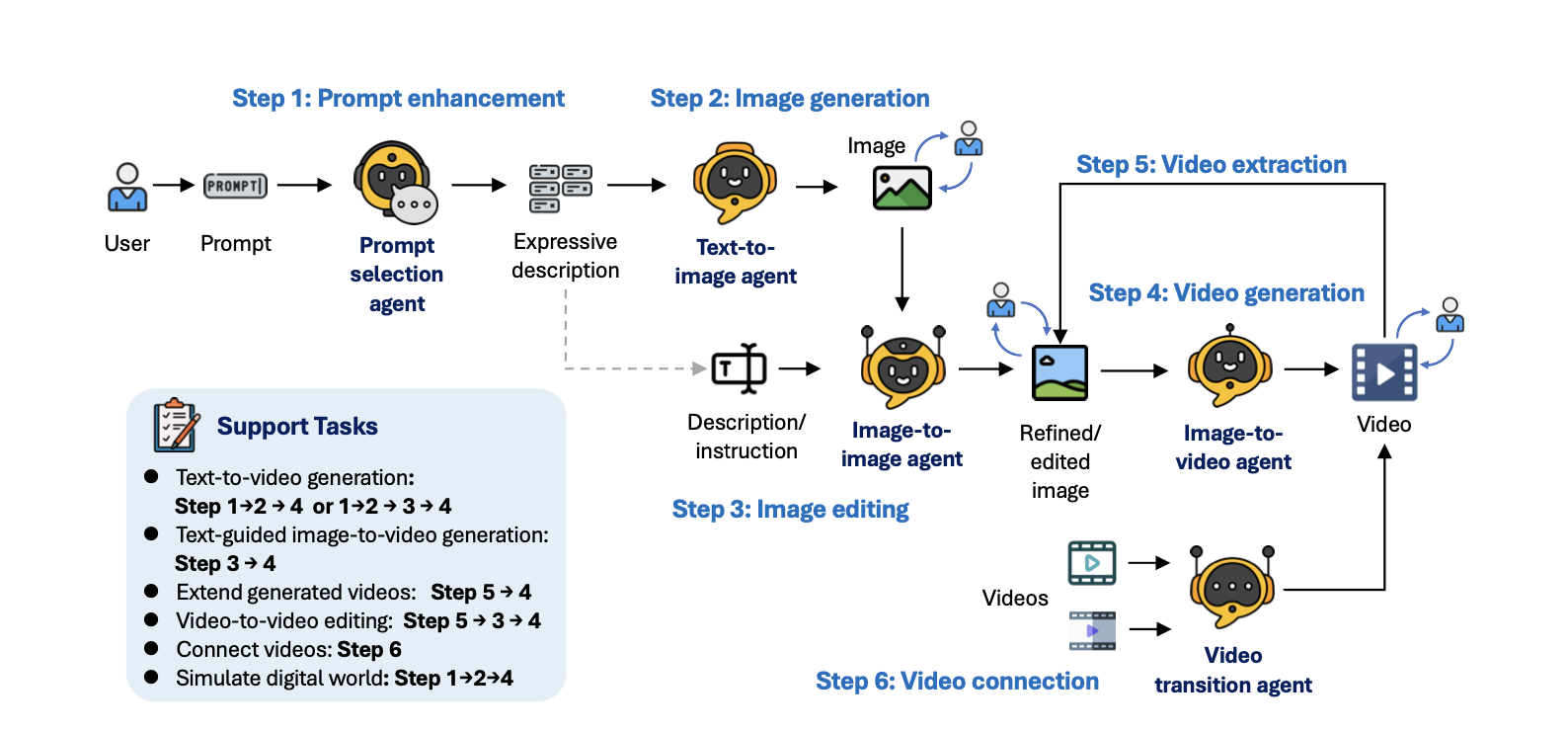}
    \caption{Mora utilizes a multi-agent framework. The prompt selection agent enhances prompts with detailed instructions, the text-to-image agent generates images from input prompts, the image-to-image agent improves the quality of photos, the text-to-video agent generates video segments, and the video transition agent integrates these videos into a longer video ~\cite{yuan2024moraenablinggeneralistvideo}.}
    \label{Multistage/Agent Based  Divide And Conquer approach}
\end{figure}

The LLM-as-Director paradigm ~\cite{lin2024videodirectorgpt}~\cite{wang2023modelscopetexttovideotechnicalreport}~\cite{10544050}~\cite{huang2023freebloomzeroshottexttovideogenerator} employs a two-stage process for video generation: (1) an LLM Planner creates narrative blueprints (keyframes, layouts, actions) and (2) a Video Generator Backbone produces intermediate frames (Fig. \ref{Video-Director-Paper}). This framework supports both zero-shot approaches like Free-Bloom ~\cite{huang2023freebloom}, which uses joint noise sampling and DDIM interpolation ~\cite{song2022denoisingdiffusionimplicitmodels}, and training-based methods like the LLM-grounded model ~\cite{lian2024llmgrounded}. VideoDirectorGPT ~\cite{lin2024videodirectorgpt} enhances zero-shot generation through GPT-4's detailed planning executed via Layout2Vid ~\cite{wang2023modelscopetexttovideotechnicalreport}, while addressing Free-Bloom's limited script utilization.

\subsubsection{Divide and Conquer Compositional/Transition  Approach  } ~\label{sec:compositional}

Early long-video generation methods stitched short clips from standard models (diffusion/autoregressive) but struggled with transitions. SEINE ~\cite{chen2023seineshorttolongvideodiffusion} innovated by framing transitions as masked diffusion, jointly denoising overlapping segments conditioned on boundary frames—though still requiring independent segment generation. Subsequent work improved continuity: MEVG ~\cite{oh2024mevgmultieventvideogeneration} anchored new clips to the final frames of predecessors, while MAVIN ~\cite{zhang2024mavinmultiactionvideogeneration} formalized transition learning as "video infilling" by training on corrupted intermediates. Encoder-Empowered GAN ~\cite{10222725} enforced temporal coherence via recall mechanisms but sacrificed dynamic content flexibility.

\textcolor{black}{These incremental advances culminated in VideoMerge~\cite{zhang2025videomergetrainingfreelongvideo}, which reimagined the paradigm entirely. Instead of post hoc stitching, it preemptively ensures coherence through (1) adaptive noise blending to unify short and long temporal scales, (2) latent fusion for boundary-free transitions, and (3) prompt refinement for persistent identity. By addressing consistency at noise, latent, and semantic levels, VideoMerge achieves what prior segment-and-merge methods could not: holistic long-video synthesis without retraining, marking a shift from compositional fixes to native long-form generation.}

The autoregressive approach ensures smooth transitions between frames by generating each frame conditioned on the previous ones, but its sequential nature makes it inherently slow for long video generation. The divide-and-conquer approach (see Section \ref{sec:llmasdirector}) can generate keyframes in parallel but faces challenges involving interpolation between frames with smooth transitions and achieving higher video quality while maintaining semantic consistency.  
Implicit generation \ref{sec:implcitvideo} approaches combine the best of both worlds by generating complete videos directly from a model conditioned on user input without the need for interpolation (divide and conquer) or extrapolation (autoregressive) between frames. A summary of key papers exploring themes related to the Divide and Conquer approach is provided in Table ~\ref{table:divideandconquercatalog}.

\begin{longtable}{ |p{3cm} |p{7cm} |c |p{2cm} |} 
\caption{Divide And Conquer Paradigms: Catalog Of key papers. In the category column,  training-free or training-based represents \ref{sec:llmasdirector} pattern
,  ’multi-stage’ represents the \ref{section:Agent-Based Divide And Conquer approach} pattern
and ’integration’ represents the \ref{sec:compositional} pattern.}
\label{table:divideandconquercatalog} \\
\hline  
 \textbf{Model}   & \textbf{Theme}  & \textbf{Month/Year}  & \textbf{category}\\ 
 \hline
\endfirsthead
\hline
\textbf{Model}   & \textbf{Theme}  & \textbf{Month/Year} & \textbf{category} \\ \hline
\endhead

\hline
\endfoot

\hline
\endlastfoot

Align your Latents~\cite{blattmann2023alignlatentshighresolutionvideo}&
Diffusion models for interpolation and upsampling.  & April 2023 &  training-based \\
   \hline  
 Gen-L-Video   ~\cite{wang2023genlvideomultitextlongvideo}&
Integrate short videos into long, consistent video.& June 2023 &   training-based \\
 \hline  

Free-Bloom ~\cite{huang2023freebloom} & LLMs and LDMs~\cite{rombach2022highresolutionimagesynthesislatent} for consistent video generation. & Sept 2023  & training-free    \\
  \hline  

VideoDirectorGPT ~\cite{lin2024videodirectorgpt}&
Consistent multi-scene videos using GPT-4 guidance. & Sept 2023  & training-free  \\
 \hline 

LVD~\cite{lian2024llmgroundedvideodiffusionmodels}&
Dynamic video scenes using LLM-guided diffusion. & Sept 2023 & training-free  \\
 \hline  

SEINE~\cite{chen2023seineshorttolongvideodiffusion}&
Long video with smooth transitions from short videos. & Oct 2023  &  integration \\
 \hline 
 Encoder GAN ~\cite{10222725}&
Connects short video by temporal relationships. & Oct 2023 &  integration \\
 \hline 
FlowZero  ~\cite{lu2023flowzero}&
Multi-frame story and aligns spatiotemporal layouts.
& Nov 2023 & training-free  \\
 \hline  
 MEVG~\cite{oh2023mtvg}& Multiple prompts, preserving visual coherence. & Dec 2023&  training-based \\  
\hline  
 
Vlogger~\cite{zhuang2024vloggermakedreamvlog} &
Specialized models to generate long videos in stages. & Jan 2024  &  multi-stage\\
 \hline  
Mora~\cite{yuan2024moraenablinggeneralistvideo}& 
Collaborative models for script, image, and video.
& March 2024 & multi-stage\\
 \hline  
  
 Vidgen ~\cite{10544050}&
LLM for story pre-processing and textual
Inversion Memory Module. & April 2024 & training-based \\
 \hline  

MAVIN~\cite{zhang2024mavinmultiactionvideogeneration}&
Transition videos creating a cohesive sequence. & May 2024  & integration\\
 \hline   

 DreamFactory~\cite{xie2024dreamfactorypioneeringmultiscenelong} &
LLM collaboration for script and movie creation.
  & August 2024 & multi-stage\\
\hline 
Kubrick~\cite{he2024kubrickmultimodalagentcollaborations}&
Agent collaborations to generate Blender scripts.  & August 2024  &  multi-stage  \\
\hline 

\textcolor{black}{VideoMerge}  ~\cite{zhang2025videomergetrainingfreelongvideo} &
\textcolor{black} {training free, merges short clips generated by pretrained text-to-video models}  & March 2025  &  multi-stage  \\
\hline 
   
\end{longtable}

\subsection{ Implicit Video Generation using compressed Latent space  }~\label{sec:implcitvideo}

\begin{figure}[H]
    \centering
    \includegraphics[scale=0.27]{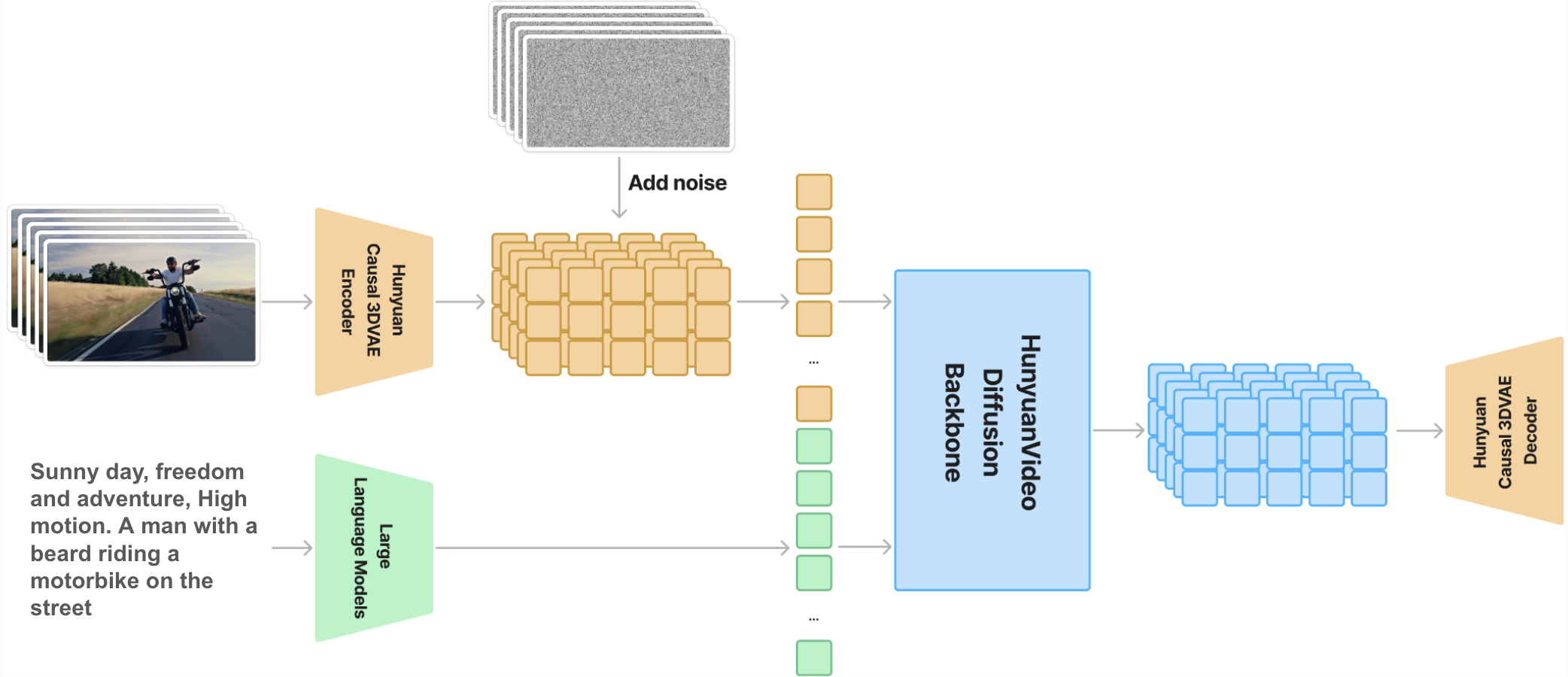}
    \caption{Using a Causal 3D VAE, Hunyuan Video compresses data into latent space. LLM-encoded text conditions Gaussian noise inputs, generating latents decoded into images/videos via the VAE decoder \cite{kong2025hunyuanvideosystematicframeworklarge}.}
    \label{Video-Generation/Hunyuan-Video}
\end{figure}

\begin{figure}[H]
    \centering
    \includegraphics[scale=0.35]{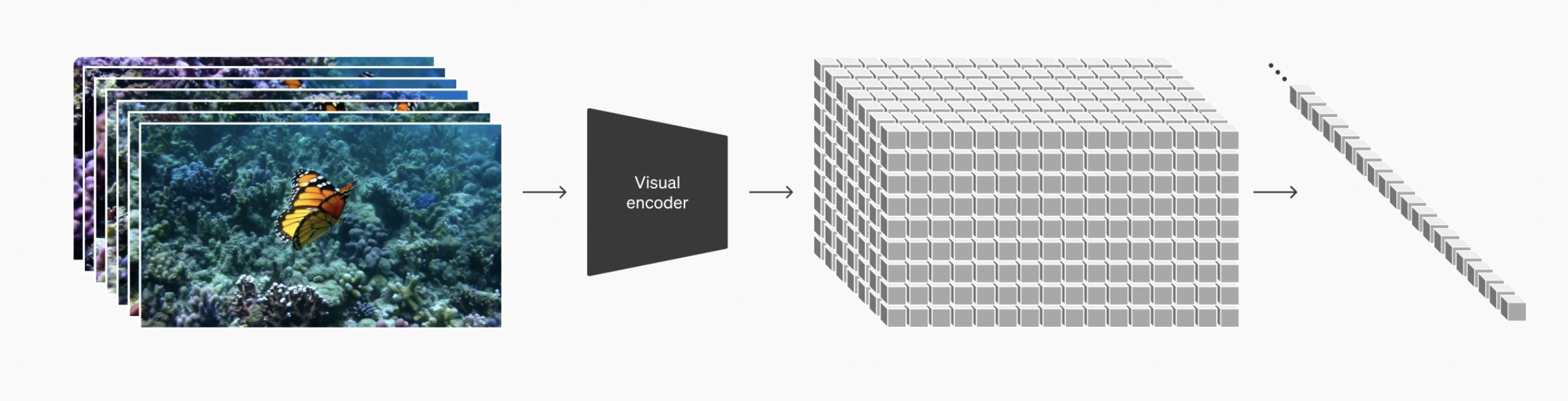}
    \caption{Transformer-Based Diffusion Model Sora compressed video of variable length into fixed space-time latent compressed representation ~\cite{soraworldmodel}.}
    \label{soraworldmodel}
\end{figure}

\textcolor{black}{Implicit video generation synthesizes complete videos simultaneously through compact latent representations, employing spacetime compression, enhanced attention mechanisms, and hierarchical denoising ~\cite{liu2024sorareviewbackgroundtechnology}. Unlike sequential approaches, models like Sora ~\cite{liu2024sorareviewbackgroundtechnology} process entire videos via: (1) spacetime compression to latent patches (visualized in Fig \ref{soraworldmodel}), (2) ViT-based denoising, and (3) LLM-augmented CLIP-like conditioning. Open-source advances include FreeNoise ~\cite{qiu2024freenoisetuningfreelongervideo} for tuning-free semantic preservation, and GLOBER ~\cite{sun2023glober} for efficient latent reconstruction. Hunyuan-Video ~\cite{kong2025hunyuanvideosystematicframeworklarge} (Fig. \ref{Video-Generation/Hunyuan-Video}) advances the field through a diffusion-VAE hybrid architecture in compressed latent space, enabling both quality and long-form coherence—demonstrating implicit generation's unique temporal synthesis capabilities from Sora to multimodal implementations.}

\textcolor{black}{ Latent-space transformer-based video generation has progressed through key architectural innovations, beginning with Goku ~\cite{chen2025gokuflowbasedvideo}'s flow-based transformers for efficient joint image-video learning. Subsequent advances include REDUCIO! ~\cite{tian2024reduciogenerating1024times1024video}'s 3D VAE compression (64× more efficient than 2D VAEs) and Mochi 1 ~\cite{genmo2024mochi}'s 10B-parameter Asymmetric Diffusion Transformer for improved coherence. Meta's Movie Gen!\cite{moviegen2024} is a unified foundation model that generates high-quality images and videos from text prompts using efficient joint training in compressed latent space. The field's current pinnacle is Cosmos ~\cite{li2025arlonboostingdiffusiontransformers}' World Foundation Model with 128× latent compression and two-phase training. However, persistent challenges in motion consistency and semantic alignment remain, evidenced by SORA's ~\cite{sora} one-minute generation limit and documented artifacts ~\cite{sun2024soraseesurveytexttovideo}. Table~\ref{table:implicit-video-generation} compares models using compressed latent spaces for video generation.}

\begin{table} [H]
    \caption{ Implicit Video Generation: Milestone Models Catalog}
    \label{table:implicit-video-generation}
    \begin{tabular}{|p{5cm} |p{8cm} |c| } 
    \hline
    \textbf{Model}  & \textbf{Theme}  & \textbf{Year}  \\
 \hline

GLOBER~\cite{sun2023glober}&
Global features to synthesize coherent video frames. & Sept 2023  \\
 \hline  
FreeNoise   ~\cite{qiu2024freenoisetuningfreelongervideo}&
Extended videos using pre-trained video diffusion models.  & Oct 2023  \\
 \hline  
 SORA    ~\cite{liu2024sorareviewbackgroundtechnology}&
Compact latent and patch-based representations. & Feb 2024  \\
 \hline  

 \textcolor{black}{Mochi 1}  ~\cite{genmo2024mochi}&
\textcolor{black} {Asymmetric Diffusion Transformer (AsymmDiT) design }& October 2024  \\
 \hline  

 \textcolor{black}{Hunyuan-Video} ~\cite{kong2025hunyuanvideosystematicframeworklarge} &
\textcolor{black} {Open source, MLLM Text Encoder and multiple video resolution
.} & Dec 2024  \\
 \hline  

 \textcolor{black}{Goku} ~\cite{chen2025gokuflowbasedvideo}&
\textcolor{black} {Flow based transformer and Vector-quantized variational autoencoder. }& Feb 2025  \\
 \hline  

  \textcolor{black}{REDUCIO!} ~\cite{tian2024reduciogenerating1024times1024video} &
\textcolor{black} { Radical latent space compression, 3D VAE that compresses videos into ultra-compact motion representations.} & Nov 2024  \\
 \hline  
 
 \textcolor{black}{Cosmos} ~\cite{nvidia2025cosmosworldfoundationmodel} &
\textcolor{black} { Physical AI based on world foundation models} & March 2025   \\
 \hline   
   
    \end{tabular}
\end{table}

\begin{figure}
    \centering
    \includegraphics[scale=0.3]{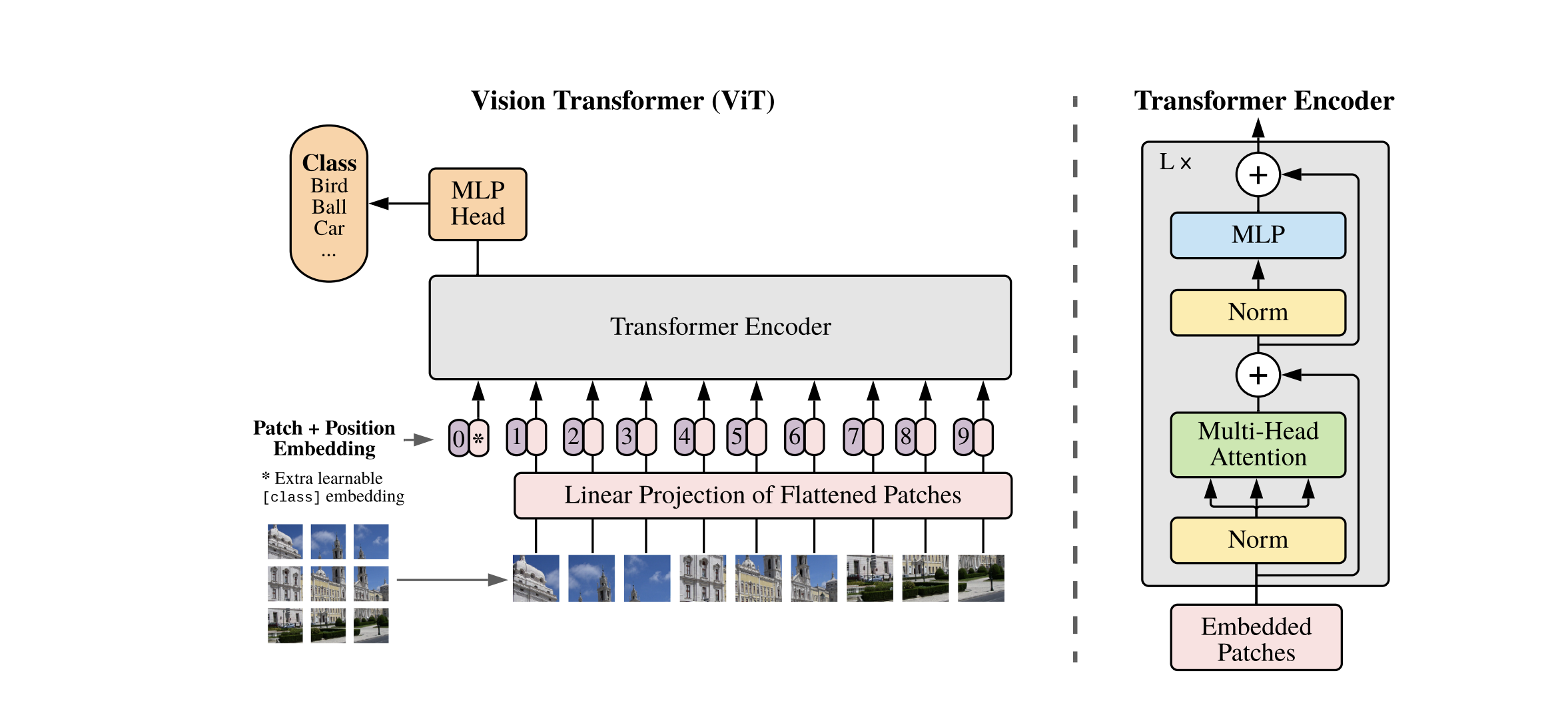}
    \caption{An image is split into fixed-size patches, linearly embedded, augmented with position embeddings, and then passed through a Transformer encoder~\cite{dosovitskiy2021imageworth16x16words}}
    \label{frametokenizationfig}
\end{figure}
Beyond generation strategies, long video modeling requires effective tokenization - representing videos as compact units for efficient processing, as we discuss in the following section.
\section{Long Video: Tokenization strategies} ~\label{sec:tokenization}
Long video tokenization strategies employ frame-level \ref{frametokenization}, temporal (3D Conv/VQ-VAE) \ref{vqvaetokenization}, and hierarchical \ref{hierarchicaltokenization} approaches to efficiently encode spatiotemporal information while balancing computational demands and representation fidelity - critical for both generative and discriminative tasks, as detailed below.
\subsection{ Frame-Level Tokenization} \label{frametokenization}
\textcolor {blue} { 
Frame-level tokenization extends Vision Transformers~\cite{dosovitskiy2021imageworth16x16words} to video via fixed-size patches (e.g., 16×16 pixels) projected into latent tokens (Fig \ref{frametokenizationfig}). Models like ViViT and MAGVIT~\cite{Yu_2023_CVPR} enhance this with spatiotemporal tokenization, hierarchical processing, and 3D embeddings. While effective, challenges remain in terms of attention complexity and motion-spatial trade-offs, driving the development of compressed token and hybrid architectures.}

\subsection {Temporal Tokenization (3D Conv/VQ-VAE)} \label{vqvaetokenization}

\begin{figure}[H]
    \centering
    \includegraphics[scale=0.3]{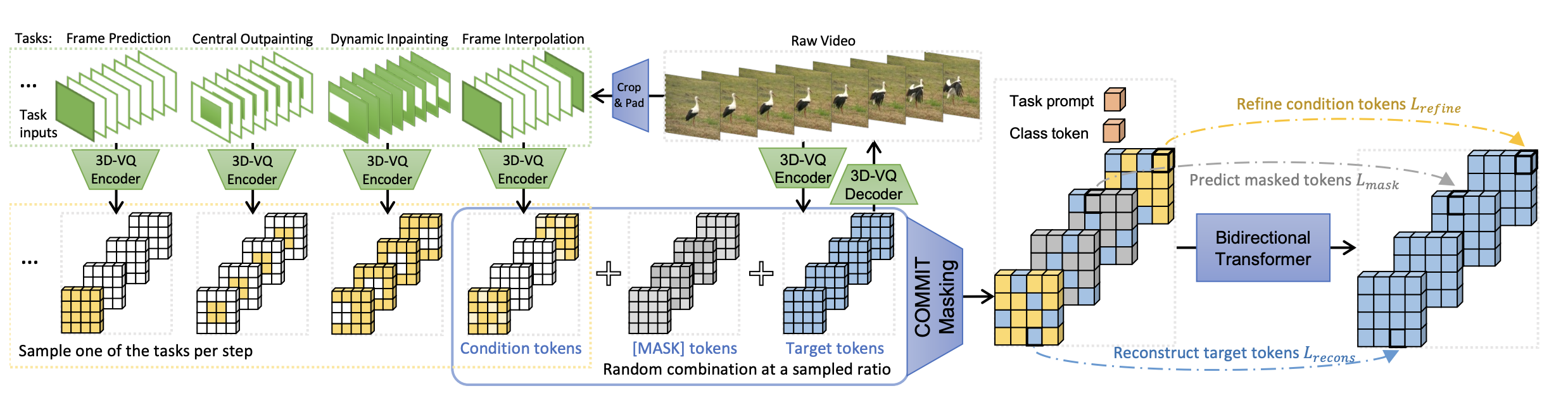}
    \caption{The proposed 3D-VQ architecture extends the 2D VQGAN \cite{esser2021tamingtransformershighresolutionimage} framework to model temporal dynamics. The system employs cascaded residual blocks in both encoder and decoder, combining 2D and 3D convolutions. ~\cite{yu2023magvitmaskedgenerativevideo}}
    \label{temporaltokenization}
\end{figure}

\textcolor{black} {Temporal tokenization compresses video into motion-aware latent representations, combining 3D convolutions with VQ-VAEs (Fig \ref{temporaltokenization})~\cite{oord2018neuraldiscreterepresentationlearning}. Pioneered by VideoGPT~\cite{yan2021videogptvideogenerationusing} and advanced in Hunyuan-Video~\cite{kong2025hunyuanvideosystematicframeworklarge} (Fig.~\ref{Video-Generation/Hunyuan-Video}), This approach tokenizes multiple spatiotemporal frames into a single compressed representation, such as MAGVIT~\cite{Yu_2023_CVPR} and cascaded approaches~\cite{ho2021cascadeddiffusionmodelshigh}, address the fidelity-efficiency tradeoffs in long-video synthesis.}

\subsection  {Hierarchical Tokenization }\label{hierarchicaltokenization}
\textcolor{black} {Hierarchical tokenization revolutionizes video compression by decomposing sequences into two optimized components: (1) high-fidelity keyframes that preserve critical visual details at sparse intervals and (2) parameterized motion representations that efficiently encode spatiotemporal dynamics between frames using compact neural descriptors. This asymmetric approach—inspired by human vision—achieves superior compression ratios while maintaining perceptual quality by decoupling spatial and temporal redundancies through learned priors, as proven in foundational work by HiTVideo~\cite{zhou2025hitvideohierarchicaltokenizersenhancing}. Recent advancements integrate attention mechanisms for adaptive keyframe selection and diffusion-based motion estimation, further enhancing the framework's efficiency and fidelity.}

In addition to the tokenization strategy, another theme in the long videos is the use of input control mechanisms, such as text, bounding boxes, and images, for video guidance. We will discuss that in the next section.
\section{Long Video: Input Control}  ~\label{sec:video-input-control}
Input conditioning involves diffusion models, GANs, or autoencoders using signals from user text prompts, entity layouts, bounding boxes, and images to condition video generation. Although long video generation utilizes many of the same techniques for input control as image and video generation models, it also faces the additional challenge of preserving long-term dependencies. Video generation models utilize innovative strategies like use of LLM  to create progressive prompts from single input prompt~\cite{huang2023freebloom}, ~\cite {hong2024direct2vlargelanguagemodels},~\cite{lian2024llmgroundedvideodiffusionmodels}, ~\cite{wang2023microcinema}, ~\cite{lin2024videodirectorgpt} and enhancement in generation mechanism to create semantic consistency between frames ~\cite{huang2023freebloom}, ~\cite{lu2023flowzero}, ~\cite {hong2024direct2vlargelanguagemodels}.

Popular mechanisms for input conditioning of long videos include User Textual Prompt, User Textual Prompt with Scene Layout, and Image Input with Textual Prompt and Scene Layout, which we will discuss next.

\subsection{User Textual Prompt} ~\label{sec:inputTextPrompt}

\begin{figure}[H]
    \centering
    \includegraphics[scale=0.38]{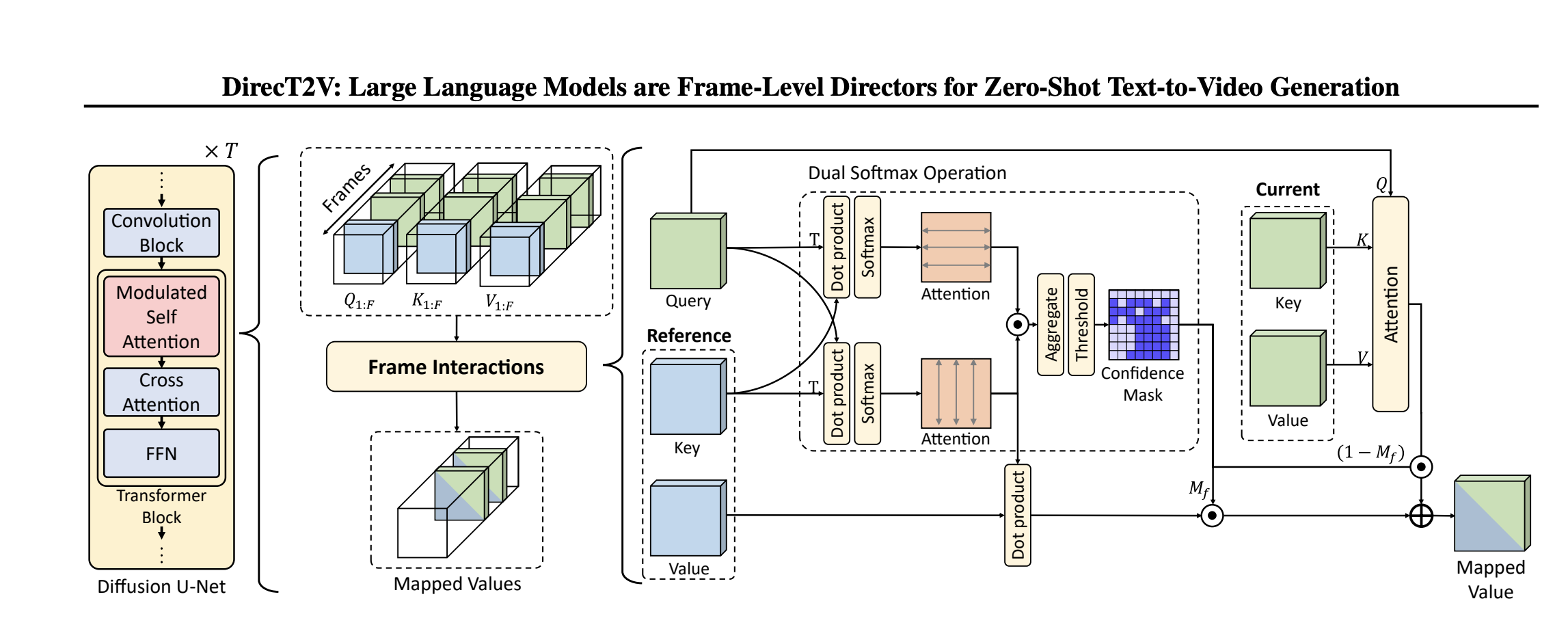}
    \caption{DirectT2V Modulated self-attention for capturing interactions between frames ~\cite{hong2024direct2vlargelanguagemodels}.}
    \label{Direct-2TV}
\end{figure}

Text prompts serve as the primary conditioning method for video generation models, with implementations ranging from single-prompt to multi-prompt approaches. Early transformer-based models, such as DALL-E ~\cite{ramesh2021zeroshottexttoimagegeneration} and CogVideo ~\cite{hong2022cogvideolargescalepretrainingtexttovideo}, utilized autoregressive transformers on joint text-image token distributions, albeit with limitations to single prompts. Phenaki ~\cite{villegas2022phenaki} advanced this by incorporating T5X embeddings ~\cite{roberts2022scalingmodelsdatatextttt5x} for sequential prompt conditioning, though facing coherence challenges in transitions. Contemporary solutions address these limitations through various approaches: Free-Bloom ~\cite{huang2023freebloom} employs LLM-generated coherent prompts with spatial-temporal attention; LLM-Grounded Video Diffusion ~\cite{lian2024llmgroundedvideodiffusionmodels} alternates between language guidance and denoising steps; VideoStudio ~\cite{long2024videodrafter} modifies cross-attention mechanisms; and DirecT2V ~\cite{hong2024direct2vlargelanguagemodels} utilizes GPT-4 for step-by-step prompt generation.
Fig. \ref{Direct-2TV} illustrates the DirecT2V architecture \cite{hong2024direct2vlargelanguagemodels}, which modifies the attention block of U-Net and incorporates modulated self-attention. Text-only prompts can guide long video generation, but they lack the semantics necessary for fine-grained control over this process. In addition to frame descriptions, adding metadata, such as bounding boxes for entities like persons and cars, as well as background information, can help generate a more accurate depiction of videos and facilitate fine alignment between text and video.

\begin{figure}[H]
    \centering
    \includegraphics[scale=0.4]{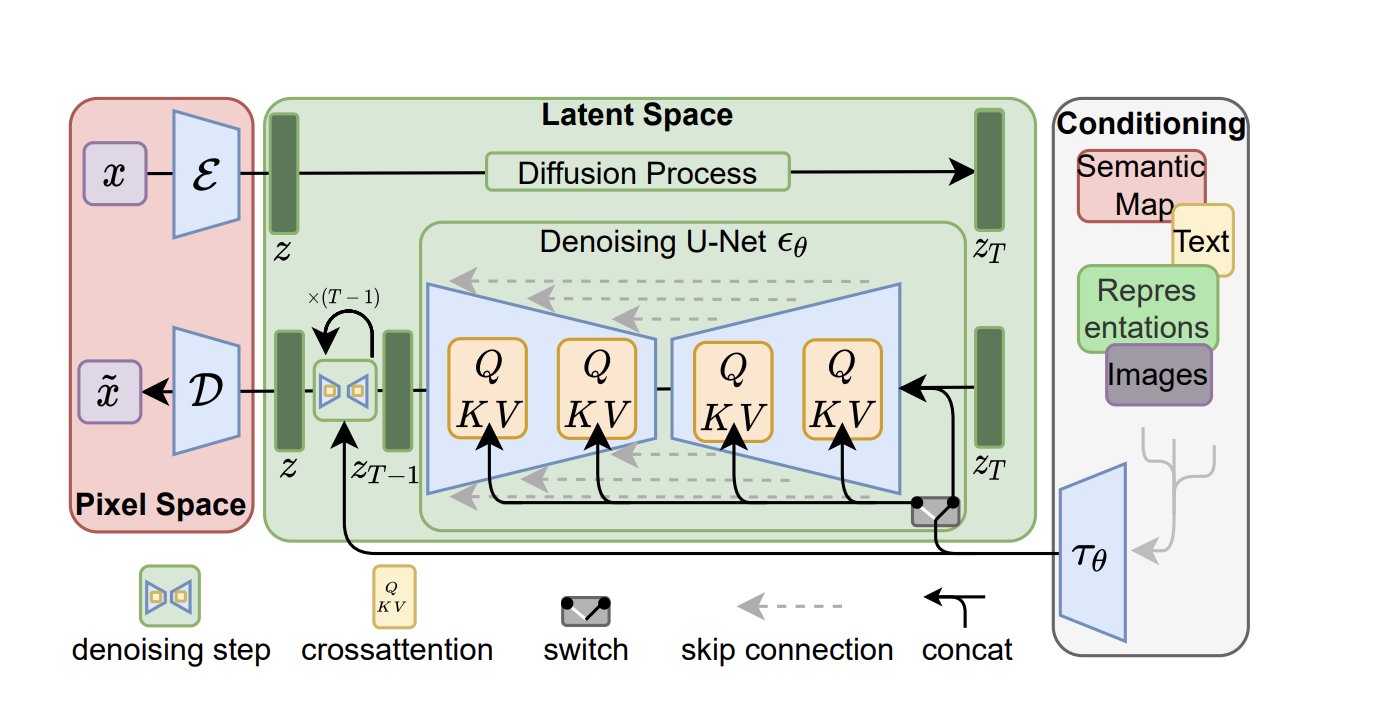}
    \caption{A latent diffusion model with input conditioning generates data by applying a reverse diffusion process on latent representations, conditioning the generation of additional input information (e.g., text or images) to guide output ~\cite{rombach2022highresolutionimagesynthesislatent}.}
    \label{llm-conditioning}
\end{figure}

\subsection{User Textual Prompt with Scenes layout} ~\label{sec:inputTextPromptWithScenes}

Multi-modal input control mechanisms extend beyond text prompts to include images, edge maps, bounding boxes, and music, thereby enhancing generation with metadata such as entity trajectories and layouts (Fig. \ref{llm-as-director}). Stable Diffusion ~\cite{rombach2022highresolutionimagesynthesislatent} pioneered this via cross-attention in UNet backbones (Fig. \ref{llm-conditioning}), while ControlNet ~\cite{zhang2023addingconditionalcontroltexttoimage} introduced a trainable copy linked via zero convolutions to preserve pre-trained features. This architecture enables diverse applications ~\cite{hu2023videocontrolnetmotionguidedvideotovideotranslation,zhang2024moonshotcontrollablevideogeneration,wang2024easycontroltransfercontrolnetvideo,zhang2023controlvideotrainingfreecontrollabletexttovideo}. Further advances include Layout2Vid in ~\cite{lin2024videodirectorgpt} (extending ModelScopeT2V ~\cite{wang2023modelscopetexttovideotechnicalreport} with layout/entity metadata), LLM-Grounded VDM’s training-free dynamic scene layouts ~\cite{lian2024llmgroundedvideodiffusionmodels}, and FlowZero’s spatiotemporal scene indices ~\cite{lu2023flowzero}. Dynamic scene layout increases control and detail alignment between text and video but does not provide aesthetic infusion. Images can guide aesthetics if generated by a high-quality text-to-image diffusion model or provided as a reference. We will discuss a few papers incorporating images as a guidance mechanism.

\subsection{Image Input  With Textual Prompt And Scenes Layout} ~\label{sec:inputTextPromptWithScenesAndImages}

Images enhance video generation by providing high-quality spatial and aesthetic details (e.g., object textures, entity positions) beyond text descriptions. NUWA-Infinity ~\cite{wu2022nuwainfinityautoregressiveautoregressivegeneration} supports both text and image inputs, while VideoStudio ~\cite{long2024videostudiogeneratingconsistentcontentmultiscene} leverages reference images (e.g., actors, objects) alongside text prompts to guide generation. Microcinema ~\cite{wang2023microcinema} employs a multi-stage pipeline, first generating images via SDXL ~\cite{podell2023sdxlimprovinglatentdiffusion} or DALL-E ~\cite{ramesh2021zeroshottexttoimagegeneration}, then using them for video synthesis. Similarly, VideoBooth ~\cite{jiang2023videoboothdiffusionbasedvideogeneration} projects reference images into CLIP ~\cite{radford2021learningtransferablevisualmodels} text space, and VideoDrafter ~\cite{long2024videodrafter} uses a two-stage control pipeline (text-to-image, then image+text-to-video). Key works are compared in Table \ref{input-control-papers-table}.

\begin{longtable}{|c |p{7cm} |p{3cm} |c|} 
\caption{ Input Conditioning: Milestone Papers. 'Text Prompt' represents \ref{sec:inputTextPrompt}, 'Text, Scene Layout' represents \ref{sec:inputTextPromptWithScenes}  and 'Image, Text, Layout' represents \ref{sec:inputTextPromptWithScenesAndImages} }
\label{input-control-papers-table} \\
\hline  
\textbf{Model}  & \textbf{Theme}  & \textbf{input control}   & \textbf{Year} \\
 \hline
\endfirsthead
\hline
\textbf{Model}  & \textbf{Theme}  & \textbf{input control}   & \textbf{Year} \\
\hline 
\endhead

\hline
\endfoot

\hline
\endlastfoot
   
Free-Bloom ~\cite{huang2023freebloom}&
LLM prompts with cross-attention and step-blocks. & Text Prompt &2023  \\
 \hline 
   
MEVG~\cite{oh2024mevgmultieventvideogeneration} &
Initial latent code, cross-attention.  & Text Prompt  &2023  \\
 \hline 
SEINE~\cite{chen2023seineshorttolongvideodiffusion}&
Transition frames using CLIP
and Lavie  ~\cite{wang2023laviehighqualityvideogeneration}.

  & Text Prompt  &2023  \\
 \hline 

 GLOBER~\cite{sun2023glober} &
CLIP encoder, cross-modal instructions with attention.
 & Text Prompt  & 2023  \\
\hline
FlowZero  ~\cite{lu2023flowzero}&
Frame sequences using LLM with cross-frame attention. & Text, Scene Layout &2023  \\
 \hline 

VideoDirectorGPT~\cite{lin2024videodirectorgpt}&
LLM generates video plan, interpreted by U-Net. & Text, Scene Layout  &2024  \\
 \hline

LLM grounded VDM ~\cite{lian2024llmgroundedvideodiffusionmodels}&
LLM story with attention maps and bounding layouts.
  &  Text, Scene Layout & 2024\\
 \hline 
 
VideoTetris~\cite{tian2024videotetriscompositionaltexttovideogeneration}&
ControlNet for autoregressive video generation. &  Text, Scene Layout &2024  \\
 \hline 

VideoDrafter~\cite{long2024videodrafter}&
LLM scripts scenes with CLIP embeddings and prompts. & Image, Text, Layout &2024  \\
 \hline 

MAVIN~\cite{zhang2024mavinmultiactionvideogeneration}&
3D UNET with cross-attention and CLIP embeddings.
 &  Image, Text, Layout  &2024  \\
 \hline 

 VideoBooth ~\cite{jiang2023videoboothdiffusionbasedvideogeneration}&
Video from images and prompts using latent space. & Image, Text, Layout & 2024  \\
 \hline 
 
MicroCinema~\cite{wang2023microcinema}&
LLM scripts multi-stage 3D Unet with cross-attention.
 &Image, Text, Layout &2024 \\
 \hline

Sora~\cite{liu2024sorareviewbackgroundtechnology} &
LLM prompts with optional visual input encoding. &Image, Text, Layout &2024  
 
\end{longtable}

User textual prompts, additional metadata for scene layouts, entity descriptions, and reference images provide a rich context for the video generation model, facilitating fine alignment between user intention and generated videos. We also require extensive video datasets with labels or captions to train the long video generation and input control mechanisms, which will be discussed in the next section.

\section{Existing Datasets}  ~\label{sec:data} 
The existing datasets for long video generation can be categorized as classification datasets \ref{sec:classification} and captions datasets \ref{sec:captions} 

\subsection{Classification Datasets} ~\label{sec:classification}

Video classification datasets have evolved significantly in scale and annotation granularity. Early datasets like UCF-101 ~\cite{soomro2012ucf101dataset101human} (13,320 clips, 101 action classes) were limited to single-label categorization. The Kinetics series ~\cite{kay2017kineticshumanactionvideo} expanded this to 306,245 videos across 700 classes while maintaining single-label classification. YouTube-8M ~\cite{abuelhaija2016youtube8mlargescalevideoclassification} introduced multi-label annotation at scale (8M videos, 350k+ hours). HowTo100M ~\cite{miech2019howto100mlearningtextvideoembedding} further advanced this with 136M clips featuring 23k tasks and narrative descriptions. The shift toward richer annotations is exemplified by captioning datasets like MSR-VTT ~\cite{7780940}, which pairs video frames with descriptive sentences. This progression reflects a broader trend from constrained single-label datasets to large-scale, multi-modal video-text collections.
 
\subsection{Captions Datasets} ~\label{sec:captions}

\begin{figure}[H]
    \centering
    \includegraphics[scale=0.7]{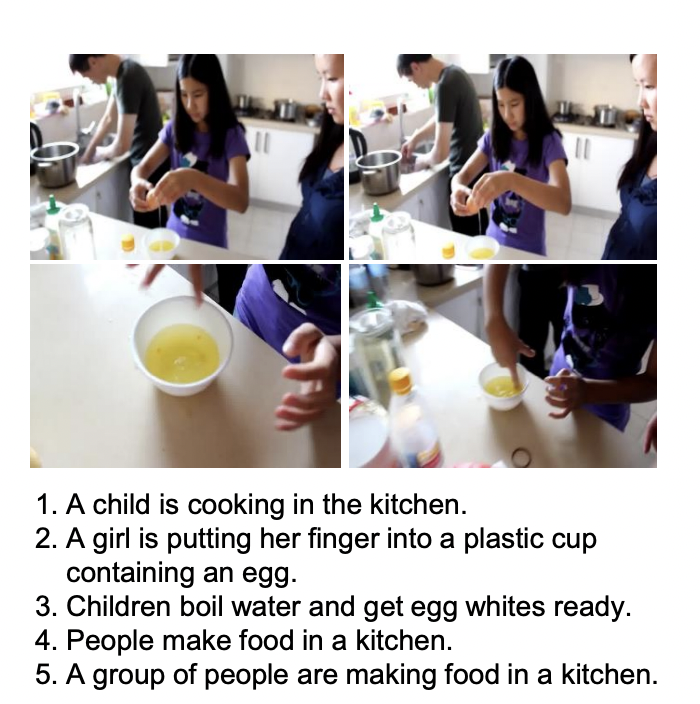}
    \caption{Here are examples from the MSR-VTT dataset showcasing video clips paired with labeled sentences. Each example includes four frames representing the video clip and five human-generated sentences that describe the content ~\cite{7780940}.}
    \label{MSR-VTT}
\end{figure}

MSR-VTT pioneered natural language video descriptions with 41.2 hours of videos and 200K clip-sentence pairs (Fig. \ref{MSR-VTT}). Later datasets dramatically scaled up volume through automated methods: WebVid-2M ~\cite{bain2022frozentimejointvideo} compiled 2M videos with algorithmic captions (similar to Conceptual Captions ~\cite{sharma2018conceptual}). In contrast, InternVid ~\cite{wang2024internvidlargescalevideotextdataset} expanded to 234M clips with 4.1B words. However, these lack detailed spatiotemporal context due to limitations in automated captioning. Recent datasets address this quality gap: VideoInstruct-100K ~\cite{maaz2024videochatgptdetailedvideounderstanding} enriched ActivityNet ~\cite{Heilbron_2015_CVPR} subsets with human-annotated spatial/temporal details, and MiraData ~\cite{ju2024miradatalargescalevideodataset} employed GPT4-V to generate structured "dense captions" covering subjects, motion, and scene attributes.

\vspace{-5pt}

Here are examples from the MSR-VTT dataset showcasing video clips paired with labeled sentences. Each example includes four frames representing the video clip and five human-generated sentences that describe the content ~\cite{7780940}. With these datasets, the next challenge is measuring the quality of generated videos, which involves assessing various aspects, such as the quality of individual frame images, the smoothness of frame transitions, and the alignment between the text and the generated video, as discussed in \ref{sec:semanticsquality}. Table ~\ref{dataset_long_video} catalogs milestone dataset used for long video generation.
 
 \begin{longtable}{ |p{4cm} |c |c| p{2cm}|p{2cm}|} 
 \caption{ Datasets For Long Videos}
\label{dataset_long_video} \\
\hline  
 \textbf{Dataset}   & \textbf{Size}  & \textbf{Month/Year} & \textbf{ Avg duration}  & \textbf{Category}\\ 
 \hline
\endfirsthead

\hline
\textbf{Dataset}   & \textbf{Size}  & \textbf{Year} & \textbf{ Avg duration}   & \textbf{Category}\\ \hline
\endhead

\hline
\endfoot

\hline
\endlastfoot
    
UCF-101 ~\cite{soomro2012ucf101dataset101human} &
13320   & Dec 2012 &  7.21 sec &     classification\\
\hline 
Kinetics-400~\cite{kay2017kineticshumanactionvideo}&
 306,245 & May 2017 & 10 sec   & classification\\
 \hline 
 Kinetics-600~\cite{carreira2018shortnotekinetics600}&
 480,000 & Aug 2018 & 10 sec  & classification\\
 \hline  

 HowTo100M  \cite{miech2019howto100mlearningtextvideoembedding} &
 136 million & June 2019  &  6.5 min    & classification \\
  \hline 

YouTube 8M  \cite{abuelhaija2016youtube8mlargescalevideoclassification} &
  6.1 million & Sept 2016  &  230 sec  & classification \\
  \hline 
   
 YouTube 8M Segments \cite{abuelhaija2016youtube8mlargescalevideoclassification}  &  237k & June 2019  &  25 sec   & classification \\
  \hline 

WebVid-2M ~\cite{bain2022frozentimejointvideo}&
2.5 million & April 2021 & 18 sec  &  captions\\
 \hline 

 Pandas 70m ~\cite{chen2024panda70mcaptioning70mvideos} &
 70.8 million & Feb 2024 & 8.5 sec  & captions\\
\hline

 HD-VG-130M ~\cite{wang2024swapattentionspatiotemporaldiffusions} &
 130 million & May 2023  &  10 sec & captions\\
\hline
InternVid ~\cite{wang2024internvidlargescalevideotextdataset} &
 234 million  & July 2023  &  39 sec   & captions\\
\hline
VidProM \cite{wang2024vidprommillionscalerealpromptgallery} & 6.69 million  & Sept 2024 & 2.5 sec  & captions\\ 
\hline 
Ego4D \cite{grauman2022ego4dworld3000hours} &
 3670(hours) & Oct 2021  &  180-300 sec 
 & captions\\
 \hline 
 
 E-SyncVidStory  \cite{yang2024synchronizedvideostorytellinggenerating} &
 6k & May 2024  &  39(s)   & captions \\
 \hline 
 LGVQ  \cite{zhang2024benchmarkingaigcvideoquality} &
 2808 & July 2024  &  8-96 sec  & captions\\
\hline
 VideoInstruct-100K \cite{maaz2024videochatgptdetailedvideounderstanding} & 100k & June 2024  &  2-3 min  & captions\\
\hline
 MiraData ~\cite {ju2024miradatalargescalevideodataset} & 788k &  July 2024 & 72.1(s)   
 & captions\\
\hline
 
 Vimeo25M \cite {wang2023laviehighqualityvideogeneration} & 25M & Sept 2023 & 19.6(s)  & captions \\

 \hline 
\end{longtable}

Long Video generation needs extensive metrics to measure aesthetic, motion, and semantic alignment between text prompts and video. We will discuss these metrics in the next section. 

\section{Performance Measures}   ~\label{sec:metrics}       
Video generation metrics can be mainly categorized into four categories, including \textit{Image Quality Metrics} \ref{sec:imagequality}, \ textit {Video Quality Metrics} \ref{sec:videoquality}, \ textit {Semantics Quality Metrics} \ref{sec:semanticsquality}, and \ textit {Composite Metrics} \ ref {sec:compositequality}.

\subsection{Image Quality Metrics } ~\label{sec:imagequality}

Image quality metrics are critical for evaluating generative models, with the Inception Score (IS) ~\cite{10.5555/3157096.3157346} being a widely adopted measure. IS uses a pre-trained Inception model ~\cite{szegedy2015rethinkinginceptionarchitecturecomputer} to assess quality (classification accuracy) and diversity (variety of classes) but fails to capture perceptual quality or generalize beyond ImageNet domains. To address these limitations, Fréchet Inception Distance (FID) ~\cite{10.5555/3295222.3295408} was introduced, comparing feature distributions between generated and authentic images for a more robust evaluation of perceptual and statistical fidelity. While these metrics excel for individual frames, they cannot assess temporal dynamics, such as motion transitions, a gap addressed by video-specific metrics.

\subsection{Video Quality Metrics} ~\label{sec:videoquality}
\begin{figure}[t]
    \centering
    \includegraphics[scale=0.3]{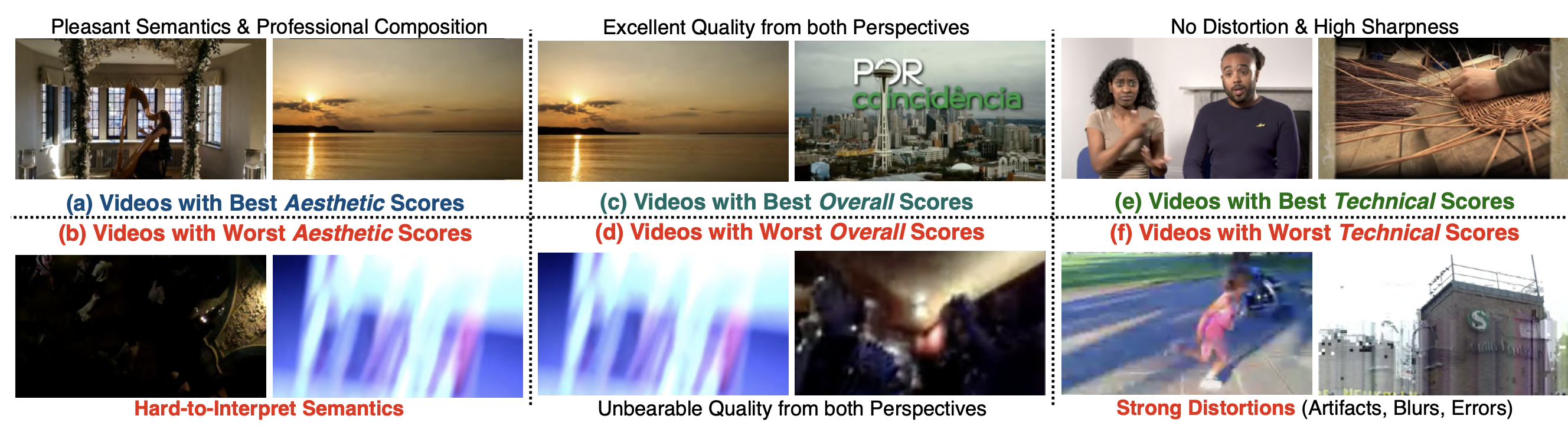}
    \caption{Dover score. Samples from a dataset with human labeling of aesthetics and technical aspects of images. Dover score could be computed by aggregating or averaging the human-assigned aesthetic and technical scores ~\cite{wu2023exploringvideoqualityassessment}.}
    \label{Dover}
\end{figure}

Video quality assessment requires metrics that evaluate both spatial and temporal coherence, extending beyond traditional image metrics such as FID. The Fréchet Video Distance (FVD) ~\cite{unterthiner2019fvd} extends FID to videos by using 3D convolutions to capture temporal dynamics (motion, transitions), though it remains computationally intensive and doesn't assess aesthetic/technical flaws. To address this, Dover ~\cite{wu2023exploringvideoqualityassessment} evaluates technical quality (blur, noise, flicker) and aesthetics, leveraging its DIVIDE-3k dataset with 450K+ subjective annotations (Fig. \ref{Dover}). For motion analysis, RAFT ~\cite{teed2020raftrecurrentallpairsfield} measures optical flow to quantify object movement and temporal alignment between frames. Together, these metrics provide complementary insights into visual fidelity, temporal coherence, and motion quality.

\vspace{-5pt} 

While video quality metrics like FVD and Dover work well for assessing the technical quality of generated videos, they have limitations in measuring how well a generated video aligns with the user's intentions or the semantic content outlined in the prompts. To address this gap, we must explore semantic alignment metrics and composite metrics, which combine technical quality and alignment with user-defined content. The metrics presented in Table \ref{video_metrics} assess how accurately the generated videos align with the intentions described in their input prompts. Additionally, Table \ref{metrics for video generation} highlights the evolution of video quality and semantic alignment metrics over time, showcasing advancements in generative models.

\subsection{Semantics Alignment Metrics } ~\label{sec:semanticsquality}

Semantic quality metrics assess how well-generated videos align with user intent, particularly in response to text prompts. CLIP ~\cite{radford2021learningtransferablevisualmodels} is a foundational tool for image-text alignment, using contrastive learning to embed both modalities into a shared space and measure their semantic similarity. Its extension, CLIPScore ~\cite{hessel2022clipscorereferencefreeevaluationmetric}, leverages CLIP’s embeddings to provide a reference-free metric for assessing image-text correspondence without ground-truth labels, making it efficient for evaluating generation models.

\textcolor{black}{CLIPSIM \cite{wu2021godivageneratingopendomainvideos} extends this paradigm for videos. CLIPSIM computes the similarity between the text and each video frame and then averages these scores to measure semantic matching.} Table \ref{tab:complete_video_metrics_v4} shows how models have been improved on these frame-level metrics and their derivatives over time.

While CLIP and CLIPScore effectively measure basic image-text alignment through embeddings, they struggle with complex object interactions or nuanced descriptions (e.g., attributes and actions). GRiT ~\cite{wu2022gritgenerativeregiontotexttransformer} addresses this by leveraging region-to-text understanding, enabling finer-grained interpretation of scenes (e.g., “a brown dog running”) and better alignment with user intent.

\begin{figure}[t]
    \centering  \includegraphics[scale=0.35]{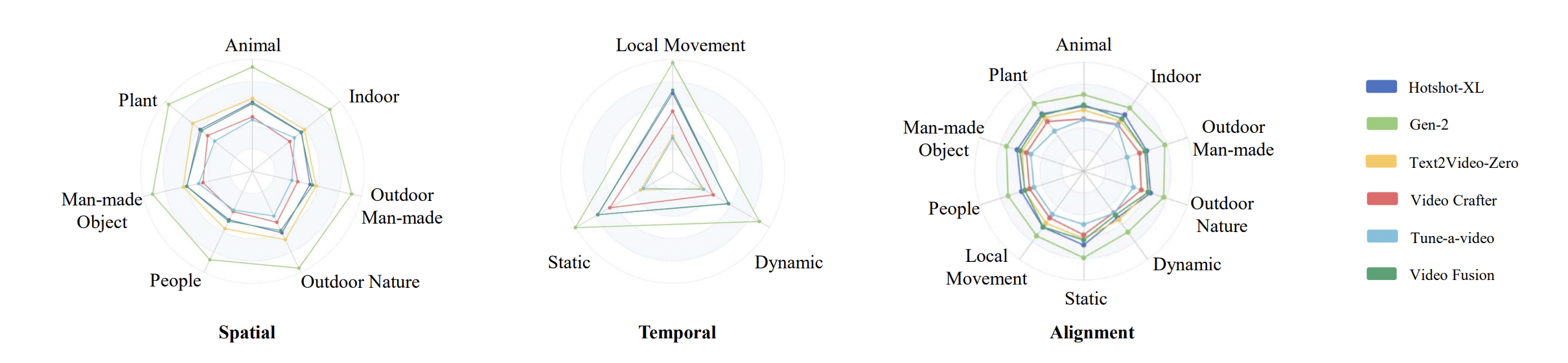}
    \caption{The prompt dataset is designed to evaluate the model by focusing on three key quality aspects: (1) spatial quality (frame appearance), (2) temporal quality (frame coherence), and (3) text-to-video alignment (content-text correspondence)~\cite{zhang2024benchmarkingaigcvideoquality}.}
    \label{UGVQ}
\end{figure}

\begin{figure}[t]
    \centering
    \includegraphics[scale=0.4]{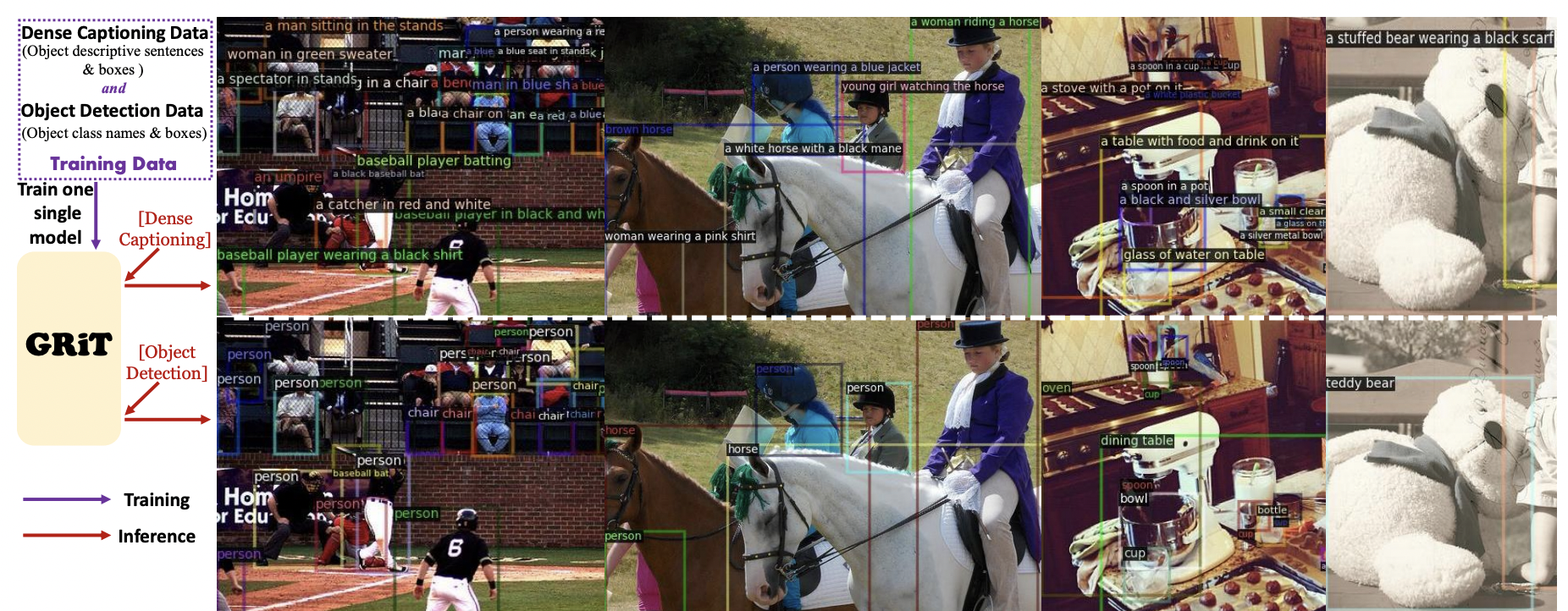}
    \caption{GRiT locates different entities in scenes with their relations and matches with dense captions ~\cite{wu2022gritgenerativeregiontotexttransformer}.}
    \label{GRIT}
\end{figure}

\begin{longtable}{|p{3cm}|c|c|c|}
\caption{Metrics for video quality evaluation. In the ``Direction'' column, $\uparrow$ represents higher score is better, while $\downarrow$ represents that a lower score is better.}
\label{video_metrics} \\ 
\hline
\textbf{Metrics} & \textbf{Type} & \textbf{Year} & \textbf{Direction} \\ 
\hline
\endfirsthead

\hline
\textbf{Metrics} & \textbf{Type} & \textbf{Year} & \textbf{Direction} \\
\hline
\endhead

\hline
\endfoot 

\hline
\endlastfoot 
 
Inception Score ~\cite{10.5555/3157096.3157346} &
frame  & June 2016   & $ \uparrow $ \\
\hline    
FID ~\cite{10.5555/3295222.3295408} &
frame  & Jan 2018  &  $ \downarrow $  \\
\hline

FVD ~\cite{unterthiner2019fvd} &video &  March 2018  &$ \downarrow $ \\
\hline

CLIPScore  ~\cite{hessel2022clipscorereferencefreeevaluationmetric}  &image  & March 2021 & $ \uparrow $ \\
\hline

FETV \cite{liu2023fetvbenchmarkfinegrainedevaluation} & video &   Nov 2023& $\uparrow $\\
\hline

VBench   ~\cite{huang2023vbenchcomprehensivebenchmarksuite} & video & Nov 2023 & $\uparrow $\\
\hline
 
 T2VQA  ~\cite{kou2024subjectivealigneddatasetmetrictexttovideo} & frame & March 2024 
  & $ \uparrow $ \\
\hline

FVD Motion ~\cite{liu2024frechetvideomotiondistance} &video & June 2024  & $ \downarrow $ \\
\hline
UGVQ ~\cite{zhang2024benchmarkingaigcvideoquality} & video & July 2024 
 &  $ \uparrow $ \\
\hline
T2V-CompBench ~\cite{sun2024t2vcompbenchcomprehensivebenchmarkcompositional} & video & July 2024 &  $\uparrow$ \\
\hline

MiraBench   ~\cite{ju2024miradatalargescalevideodataset} & video & July 2024   & $\uparrow $\\
\hline
Cross-Scene Face/Style Consistency Score  ~\cite{xie2024dreamfactorypioneeringmultiscenelong}  & video & August 2024 & $\uparrow $ \\
\end{longtable}

As illustrated in Fig. \ref{GRIT}, GRiT employs a transformer-based architecture to learn the relationships between different image regions and their corresponding textual descriptions. It enables the model to break down the image into distinct regions and understand how each part corresponds to specific components of the prompt. In conclusion, while CLIP and CLIPScore provide practical methods for measuring the similarity between images and text, GRiT offers a significant advancement by enabling a deeper semantic understanding of the content within images. By considering not only individual objects but also their relationships, attributes, and actions, GRiT enhances the ability to evaluate generated content on a much more complex and nuanced level. These advances in semantic quality metrics are crucial for enhancing the alignment of videos generated with user intentions, ensuring that the videos are both visually accurate and semantically meaningful.

\begin{longtable}{lccccccc}
    
    \caption{\textcolor{black}{Comprehensive benchmark comparison of video generation models ordered by publication year (2017-2024) ~\cite{guo2023videopoet}, ~\cite{lin2024videodirectorgpt}, ~\cite{singer2022makeavideo}, \cite{huang2023freebloom},\cite{oh2024mevgmultieventvideogeneration}, \cite{lu2023flowzero}, \cite{lian2024llmgrounded}, \cite{he2023latentvideodiffusionmodels}, \cite{li2023vlogger}, \cite{microcinema}, \cite{chen2023videodirectorgpt},  \cite{sun2023glober}, \cite{wang2023laviehighqualityvideogeneration},  \cite{ho2022videodiffusionmodels}, \cite{jiang2023videoboothdiffusionbasedvideogeneration}, \cite{qiu2024freenoisetuningfreelongervideo}, \cite{chen2025gokuflowbasedvideo}, \cite{tian2024reduciogenerating1024times1024video}, \cite{he2024kubrickmultimodalagentcollaborations}, \cite{xie2024dreamfactorypioneeringmultiscenelong}  }}
    \label{tab:complete_video_metrics_v4}\\
    \toprule
    \label{metrics for video generation} \\
    \textbf{Model (Citation)} & \textbf{Year} & \textbf{Zero-Shot} & \textbf{Samples} & \textbf{FVD} & \textbf{FID} & \textbf{CLIPSIM} & \textbf{IS} \\ 
    \midrule
    \endfirsthead
    
    \multicolumn{8}{c}%
    {{\bfseries \tablename\ \thetable{} -- continued from previous page}} \\
    \toprule
    \textbf{Model (Citation)} & \textbf{Year} & \textbf{Zero-Shot} & \textbf{Samples} & \textbf{FVD} & \textbf{FID} & \textbf{CLIPSIM} & \textbf{IS} \\ 
    \midrule
    \endhead
    
    \bottomrule
    \multicolumn{8}{r}{{Continued on next page}} \\
    \endfoot
    
    \bottomrule
    \multicolumn{8}{l}{\footnotesize
    \begin{tabular}{ll}
    FVD: UCF-101 metrics (from Video LDM comparison table) & FID/CLIPSIM: MSR-VTT metrics \\
    BAIR FVD scores shown in FVD column & X: Not applicable, Yes: Zero-shot capable \\
    \textsuperscript{†} ModelScope replication results & Bold: Best results in each category \\
    \end{tabular}}
    \endlastfoot
    \multicolumn{8}{l}{\textit{2021 Models}} \\
    GODIVA \cite{wu2021godivageneratingopendomainvideos} & 2021 & No & 30 & -- & -- & 0.2402 & -- \\
    NUWA \cite{wu2021nuwa} & 2021 & X & 0.97M & -- & 28.46 & -- & -- \\
    VideoGPT \cite{yan2021videogptvideogenerationusing} & 2021 & -- & -- & 103.3 & 24.69 & -- & 24.69 \\
    Video Transformer \cite{arnab2021vivitvideovisiontransformer} & 2021 & -- & -- & 94 ± 2 & -- & -- & -- \\
    TGAN-F \cite{saito2017temporalgenerativeadversarialnets} & 2021 & -- & -- & -- & 7817 & -- & 22.91 \\
    LVT \cite{huang2021lvt} & 2021 & -- & -- & 125.8 & -- & -- & -- \\
    VGAN \cite{vondrick2016vgan} & 2021 & -- & -- & -- & -- & -- & 8.31 \\
    
    \multicolumn{8}{l}{\textit{2022 Models}} \\
    Make-A-Video \cite{singer2022makeavideo} & 2022 & Yes & 1 & 367.23 & 13.17 & 0.3049 & 33.00 \\
    CogVideo (Chinese) \cite{hong2022cogvideo} & 2022 & Yes & 1 & 701.59 & 23.59 & 0.2614 & 23.55 \\
    CogVideo (English) \cite{hong2022cogvideo} & 2022 & Yes & 1 & 701.59 & -- & 0.2631 & 25.27 \\
    VideoFusion \cite{luo2023videofusion} & 2022 & -- & -- & 639.90 & -- & -- & -- \\
    LVDM \cite{he2023latentvideodiffusionmodels} & 2022 & -- & -- & 372.00 & -- & -- & -- \\
    Video Diffusion \cite{ho2022video} & 2022 & -- & -- & -- & 295 & -- & 57 \\
    Phenaki \cite{villegas2022phenaki} & 2022 & Yes & 15M & -- & 37.74 & -- & -- \\
    MagicVideo \cite{zhao2023magicvideo} & 2022 & -- & -- & 655.00 & -- & -- & -- \\
    
    \multicolumn{8}{l}{\textit{2023 Models}} \\
    PYoCo \cite{blattmann2023pyoco} & 2023 & -- & -- & 355.19 & -- & \textbf{0.3204} & \textbf{47.76} \\
    VideoPoet (Pretrain) \cite{guo2023videopoet} & 2023 & -- & -- & 355 & -- & 0.3049 & 38.44 \\
    VideoFactory \cite{wang2023videofactory} & 2023 & -- & -- & 410.00 & -- & 0.3005 & -- \\
    ModelScope \cite{chen2023modelscope} & 2023 & -- & -- & 410.00 & 12.32 & 0.2930 & -- \\
    LaViE \cite{ma2023lavie} & 2023 & -- & -- & 526.30 & -- & 0.2949 & -- \\
    Video LDM \cite{blattmann2023videoldm} & 2023 & -- & -- & 550.61 & -- & 0.2929 & 33.45 \\
    Vlogger \cite{li2023vlogger} & 2023 & Yes & 10M & 292.43 & 37.23 & -- & -- \\
    ModelScopeT2V \cite{chen2023modelscope} & 2023 & -- & -- & -- & 12.32 & 0.2909 & -- \\
    VideoDirectorGPT \cite{chen2023videodirectorgpt} & 2023 & -- & -- & -- & 12.22 & 0.2860 & -- \\
    InternVid \cite{wang2023internvid} & 2023 & -- & -- & 617 & -- & 0.2951 & 21.04 \\
    MicroCinema \cite{microcinema} & 2023 & -- & -- & 342.86 & -- & -- & -- \\
    PixelDance \cite{zeng2023makepixelsdancehighdynamic} & 2023 & -- & -- & 242.82 & -- & -- & 42.10 \\
    Emu-Video \cite{girdhar2024emuvideofactorizingtexttovideo} & 2023 & -- & -- & 317.10 & -- & -- & 42.70 \\
    \multicolumn{8}{l}{\textit{2024 Models}} \\
    Lumiere \cite{lumiere} & 2024 & -- & -- & 332.49 & -- & -- & -- \\
    Reducio-DiT \cite{tian2024reduciogenerating1024times1024video} & 2024 & -- & -- & 318.50 & -- & -- & -- \\
    Goku-2B (256×256)\cite{chen2025gokuflowbasedvideo} & 2024 & Yes & -- & 246.17 & -- & -- & 45.77 ± 1.10 \\
    \end{longtable}

These semantic alignment metrics have limitations in the video domain because the video may contain hundreds of frames, and they must match caption boundaries with corresponding frames. Composite metrics, aggregates of many individual algorithms, and manual scoring address this limitation. 

\begin{figure}[H]
    \centering   \includegraphics[scale=0.35]{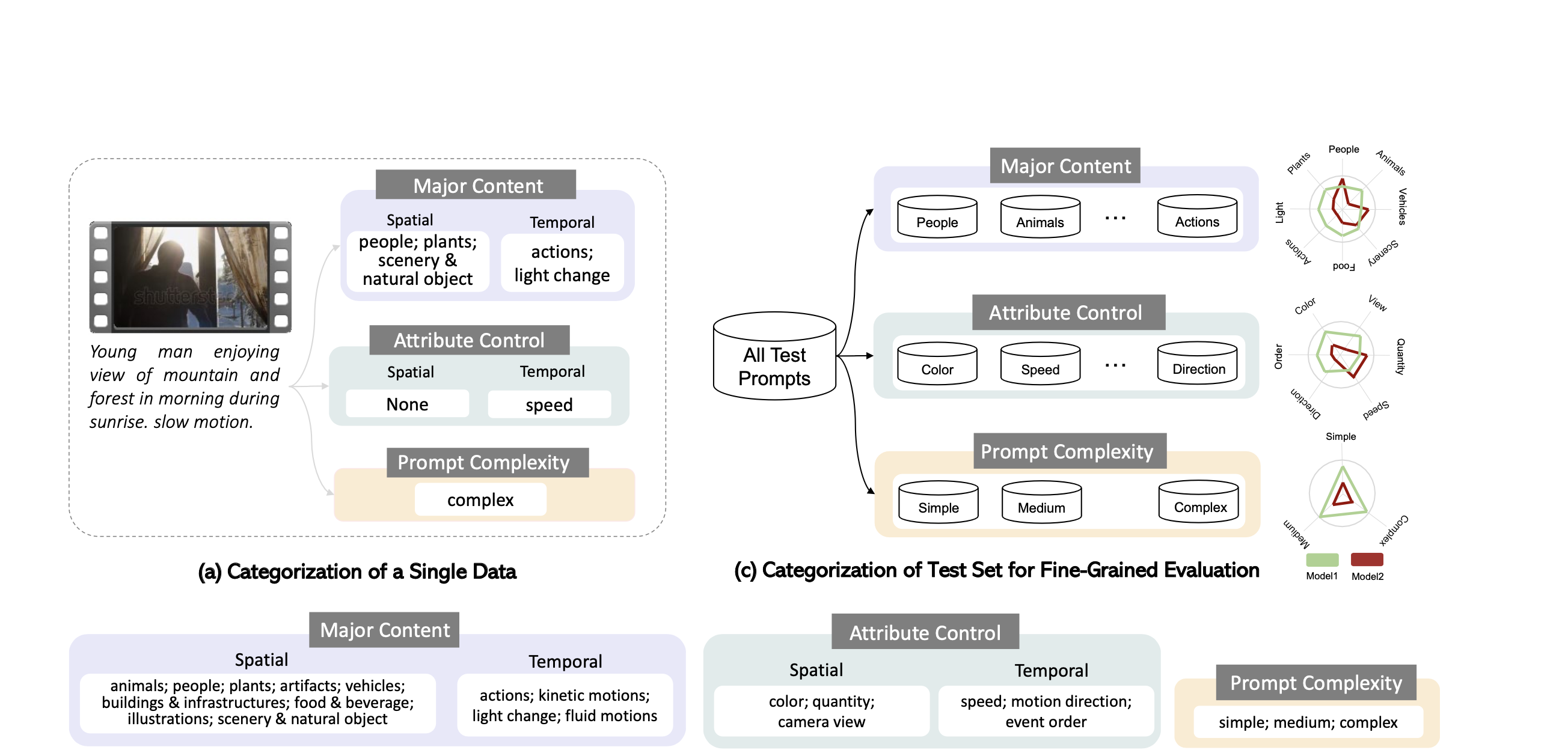}
    \caption{FETV is multi-faceted, classifying prompts into three distinct aspects: the main content, controllable attributes, and prompt complexity \cite{liu2023fetvbenchmarkfinegrainedevaluation}.
}
    \label{FETV}
\end{figure}

\subsection{Composite Metrics} ~\label{sec:compositequality}

Composite metrics combine multiple evaluation approaches to assess text-to-video generation across diverse categories (animals, objects, people) and dimensions (spatial, temporal, motion alignment). FETV Bench ~\cite{liu2023fetvbenchmarkfinegrainedevaluation} pioneered this approach with its three-aspect prompt system (content, attributes, complexity Fig \ref{FETV}) and hybrid evaluation using both manual labeling and automated metrics like CLIPScore ~\cite{hessel2022clipscorereferencefreeevaluationmetric} and BLIPScore ~\cite{lu2023llmscoreunveilingpowerlarge}. UGVQ ~\cite{zhang2024benchmarkingaigcvideoquality} expanded this framework through its LGVQ dataset, evaluating spatial quality, motion coherence, and text-video alignment with partitioned prompts (foreground/background/motion) and manual scoring of six models (Fig. \ref{UGVQ}). Subsequent benchmarks, such as T2V-CompBench, VBench/VBench++, and MiraBench, have further developed comprehensive evaluation protocols, although they reveal persistent gaps in the reliability of automated metrics - particularly for assessing long video temporal consistency and semantic fidelity (Table \ref{FETVMetrics}).

\begin{longtable}{@{}p{2.5cm} p{1.3cm} p{1.3cm} p{1.3cm} p{0.9cm} p{0.9cm} p{0.9cm}@{}}
\caption{Comparative analysis of video generation models across 12 VBench dimensions 
\label{FETVMetrics}
\cite{liu2023fetvbenchmarkfinegrainedevaluation,huang2023vbenchcomprehensivebenchmarksuite}}\label{tab:vbench}\\
\toprule
\multicolumn{1}{c}{\textbf{\scriptsize Models}} & 
\multicolumn{1}{c}{\textbf{\scriptsize Subj. Cons.}} & 
\multicolumn{1}{c}{\textbf{\scriptsize Bkg. Cons.}} & 
\multicolumn{1}{c}{\textbf{\scriptsize Temp. Flicker}} & 
\multicolumn{1}{c}{\textbf{\scriptsize Motion}} & 
\multicolumn{1}{c}{\textbf{\scriptsize Aesthetic}} & 
\multicolumn{1}{c}{\textbf{\scriptsize Obj. Class}} \\
\midrule
\endfirsthead

\caption[]{Video generation model comparison (continued)}\\
\toprule
\multicolumn{1}{c}{\textbf{\scriptsize Models}} & 
\multicolumn{1}{c}{\textbf{\scriptsize Subj. Cons.}} & 
\multicolumn{1}{c}{\textbf{\scriptsize Bkg. Cons.}} & 
\multicolumn{1}{c}{\textbf{\scriptsize Temp. Flicker}} & 
\multicolumn{1}{c}{\textbf{\scriptsize Motion}} & 
\multicolumn{1}{c}{\textbf{\scriptsize Aesthetic}} & 
\multicolumn{1}{c}{\textbf{\scriptsize Obj. Class}} \\
\midrule
\endhead

\bottomrule
\multicolumn{7}{r@{}}{Continued on next page} \\
\endfoot
\bottomrule
\endlastfoot

LaVie~\cite{wang2023lavie} & 91.41 & 97.47 & 98.30 & 96.38 & 54.94 & 91.82 \\
ModelScope~\cite{wang2023modelscope} & 89.87 & 95.29 & 98.28 & 95.79 & 52.06 & 82.25 \\
CogVideo~\cite{hong2022cogvideo} & 92.19 & 96.20 & 97.64 & 96.47 & 38.18 & 73.40 \\
VideoCrafter~\cite{chen2024videocrafter2overcomingdatalimitations} & 96.85 & 98.22 & 98.41 & 97.73 & 63.13 & 92.55 \\
Gen-2~\cite{Gen3} & 97.61 & 97.61 & 99.56 & 99.58 & 66.96 & 90.92 \\
AnimateDiff~\cite{guo2023animatediff} & 95.30 & 97.68 & 98.75 & 97.76 & 67.16 & 90.90 \\
Latte-1~\cite{ma2024latte} & 88.88 & 95.40 & 98.89 & 94.63 & 61.59 & 86.53 \\
Pika-1.0~\cite{pikalab} & 96.94 & 97.36 & 99.74 & 99.50 & 62.04 & 88.72 \\
Kling~\cite{kling} & 98.33 & 97.60 & 99.30 & 99.40 & 61.21 & 87.24 \\
Gen-3~\cite{Gen3} & 97.10 & 96.62 & 98.61 & 99.23 & 63.34 & 87.81 \\
CogVideoX~\cite{yang2024cogvideox} & 96.23 & 96.52 & 98.66 & 96.92 & 61.98 & 85.23 \\

\midrule
\multicolumn{1}{c}{\textbf{\scriptsize Models}} & 
\multicolumn{1}{c}{\textbf{\scriptsize Mult. Obj.}} & 
\multicolumn{1}{c}{\textbf{\scriptsize Human Act.}} & 
\multicolumn{1}{c}{\textbf{\scriptsize Color}} & 
\multicolumn{1}{c}{\textbf{\scriptsize Spatial}} & 
\multicolumn{1}{c}{\textbf{\scriptsize Temp. Style}} & 
\multicolumn{1}{c}{\textbf{\scriptsize Overall}} \\
\midrule

LaVie~\cite{wang2023lavie} & 33.32 & 96.80 & 86.39 & 34.09 & 25.93 & 26.41 \\
ModelScope~\cite{wang2023modelscope} & 38.98 & 92.40 & 81.72 & 33.68 & 25.37 & 25.67 \\
CogVideo~\cite{hong2022cogvideo} & 18.11 & 78.20 & 79.57 & 18.24 & 7.80 & 7.70 \\
VideoCrafter~\cite{chen2024videocrafter2overcomingdatalimitations} & 40.66 & 95.00 & 92.92 & 35.86 & 25.84 & 28.23 \\
Gen-2~\cite{wang2023lavie} & 55.47 & 89.20 & 89.49 & 66.91 & 24.12 & 26.17 \\
AnimateDiff~\cite{guo2023animatediff} & 36.88 & 92.60 & 87.47 & 34.60 & 26.03 & 27.04 \\
Latte-1~\cite{ma2024latte} & 34.53 & 90.00 & 85.31 & 41.53 & 24.76 & 27.33 \\
Pika-1.0~\cite{pikalab} & 43.08 & 86.20 & 90.57 & 61.03 & 24.22 & 25.94 \\
Kling~\cite{kling} & 68.05 & 93.40 & 89.90 & 73.03 & 24.17 & 26.42 \\
Gen-3~\cite{Gen3} & 53.64 & 96.40 & 80.90 & 65.09 & 24.71 & 26.69 \\
CogVideoX~\cite{yang2024cogvideox} & 62.11 & 99.40 & 82.81 & 66.35 & 25.38 & 27.59 \\
\end{longtable}

\section{Conclusion \& Future Trends}  
This survey equips users with a broad overview of the history, recent progress, and ongoing challenges in long video generation, focusing on video generation strategies, datasets, metrics, and open research areas. Long video generation is one of the actual north goals of generative AI, aiming to produce coherent and realistic videos over extended durations. Some of the challenges to be addressed by long video generation are maintaining temporal coherence and visual consistency while ensuring that the generated video aligns with a narrative or specific user intentions. Several strategies have been explored to tackle this challenge, including divide-and-conquer autoregressive models and intrinsic methods. Despite progress in these areas, motion consistency, semantic alignment, and parallel processing remain key obstacles to achieving scalable, high-quality long video generation. Future research in long video generation can focus on enhanced autoregressive models,  novel frame and video segment merging techniques, and enhanced training paradigms. 

One of the significant open research areas in long video generation is the generation of longer videos that accurately reflect spatial, temporal, and physical dynamics. A key challenge is the need for large-scale video datasets with comprehensive spatial, temporal, and physical context (e.g., trajectories, shadows, interactions). Existing large-scale datasets, such as HD-VG-130M ~\cite{wang2024swapattentionspatiotemporaldiffusions}, offer scale but have limitations in terms of caption quality (e.g., captions are restricted to 15-20 words and lack rich spatial and temporal contextual information). On the other hand, datasets such as VideoInstruct-100K ~\cite{maaz2024videochatgptdetailedvideounderstanding} provide rich spatial and temporal context but fall short in scale. Developing datasets that balance both scale and rich context is critical for advancing long video generation research. In addition to datasets, measuring the quality of generated videos presents another challenge. Current state-of-the-art metrics, such as FETV ~\cite{liu2023fetvbenchmarkfinegrainedevaluation}, MiraBench  ~\cite{ju2024miradatalargescalevideodataset}, and VBench~\cite{huang2023vbenchcomprehensivebenchmarksuite}, rely on manual human feedback to assess video quality, which is time-consuming, subjective, and challenging to scale. Future research should focus on developing fully automated metrics that can objectively evaluate the quality of generated videos in a more scalable manner.

Another open area of research in long video generation is the integration of audio. Currently, most commercial video generation models, such as SORA and Stability AI, do not produce accompanying audio. Developing methods to generate audio that aligns seamlessly with visual content is crucial for creating immersive and comprehensive videos, making this a key focus in the field of long-form video generation.

Long video generation promises to revolutionize multiple fields, including entertainment, education, virtual reality, and game development. However, it also introduces significant challenges, such as the potential for fake video creation, bias, violence, and moral concerns.   Additionally, issues like hallucinations can limit the applicability of generative videos, particularly in domains like education and science. In conclusion, this survey provides readers with an in-depth overview of the current state of the art in long video generation, highlighting key research areas and opportunities for future exploration.

\begin{center}
\href{https://longvideogeneration.s3.us-east-2.amazonaws.com/LongVideo.html}{Long Video Generation Videos Home }
\end{center}

\bibliographystyle{ACM-Reference-Format}
\bibliography{video-survey}

\appendix

\end{document}